\documentclass[conference]{IEEEtran}
\usepackage{times}

\usepackage[numbers]{natbib}
\usepackage{multicol}
\usepackage{subcaption}

\newcommand{\ie}{\emph{i.e.,~}}

\usepackage{graphicx}
\usepackage{mathrsfs}
\usepackage{amsfonts}
\usepackage{amsmath}
\usepackage{colortbl}
\usepackage{url}            
\usepackage{booktabs}       
\usepackage{amsfonts}       
\usepackage{nicefrac}       
\usepackage{microtype}  
\usepackage{marvosym}
\usepackage{xcolor}    
\usepackage{array} 
\usepackage{epsfig}
\usepackage{graphicx}
\usepackage{amsmath}
\usepackage{amssymb}
\usepackage{multirow}
\usepackage{chemformula}
\usepackage{booktabs}
\usepackage{url}
\usepackage{graphicx}
\usepackage{natbib}
\setcitestyle{numbers,square}
\usepackage{algorithm}  
\usepackage{algorithmicx} 
\usepackage{algpseudocode}
\usepackage{adjustbox}
\usepackage{caption}
\usepackage{chemformula}
\usepackage{subfloat}
\usepackage{chemformula}
\usepackage{tabularx} 
\usepackage{ragged2e} 
\usepackage{booktabs}
\usepackage{makecell}
\usepackage{wrapfig}
\usepackage{amsfonts}
\usepackage{xspace}
\usepackage{anyfontsize}
\usepackage{nicefrac}  
\usepackage[accsupp]{axessibility} 
\definecolor{darkred}{rgb}{0.6,0,0} 
\usepackage[breaklinks=true,
            colorlinks,
            linkcolor = darkred,
            urlcolor  = magenta, 
            citecolor = teal,
            bookmarks = true]{hyperref}
\begin{document}
\pagestyle{plain}
\title{CordViP: Correspondence-based Visuomotor Policy for Dexterous Manipulation in Real-World}


\author{
    \IEEEauthorblockN{
        Yankai Fu\textsuperscript{1}$^{\dagger}$, 
        Qiuxuan Feng\textsuperscript{1}$^{\dagger}$, 
        Ning Chen\textsuperscript{1}$^{\dagger}$, 
        Zichen Zhou\textsuperscript{1}, 
        Mengzhen Liu\textsuperscript{1}, \\
        Mingdong Wu\textsuperscript{1}, 
        Tianxing Chen\textsuperscript{2}, 
        Shanyu Rong\textsuperscript{1},
        Jiaming Liu\textsuperscript{1}, 
        Hao Dong\textsuperscript{1}, 
        Shanghang Zhang\textsuperscript{1,3}\textsuperscript{\Letter}
    }
    \vspace{0.5em}
    \IEEEauthorblockA{\textsuperscript{1}State Key Laboratory of Multimedia Information Processing, School of Computer Science, Peking University,}
    \IEEEauthorblockA{\textsuperscript{2}The University of Hong Kong, \textsuperscript{3}Beijing Academy of Artificial Intelligence}
    \IEEEauthorblockA{$^{\dagger}$Equal Contribution, \textsuperscript{\Letter}Corresponding author}
    \\ \normalsize{\textcolor{purple}{\url{https://aureleopku.github.io/CordViP}}} 
    \vspace{-15pt} 
}



%


\twocolumn[{
\renewcommand\twocolumn[1][]{#1}
\maketitle
\begin{center}
    \captionsetup{type=figure}
    \includegraphics[width=\textwidth]{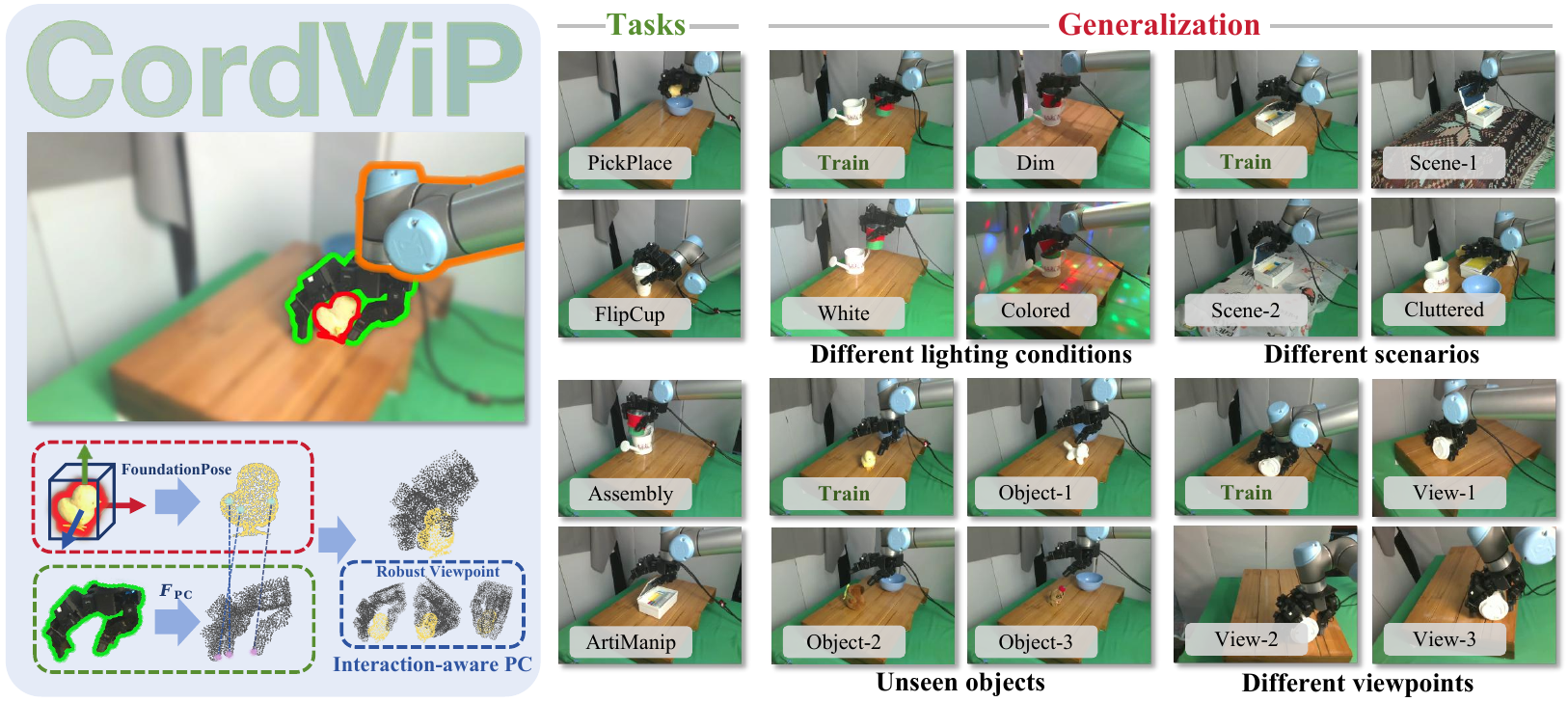}
    \captionof{figure}{We propose CordViP, a correspondence-based visuomotor policy for dexterous manipulation in the real world. (a) Left: We present the interaction-aware point clouds, which demonstrate robustness to different viewpoints while establishing correspondences between the object and the hand. (b) Right: Our method achieves promising results across multiple real-world dexterous manipulation tasks, showcasing exceptional generalization capabilities.}
\end{center}
}]


\begin{abstract}


Achieving human-level dexterity in robots is a key objective in the field of robotic manipulation. Recent advancements in 3D-based imitation learning have shown promising results, providing an effective pathway to achieve this goal.
However, obtaining high-quality 3D representations presents two key problems: 
(1) the quality of point clouds captured by a single-view camera is significantly affected by factors such as camera resolution, positioning, and occlusions caused by the dexterous hand; (2) the global point clouds lack crucial contact information and spatial correspondences, which are necessary for fine-grained dexterous manipulation tasks.
To eliminate these limitations, we propose CordViP, a novel framework that constructs and learns correspondences by leveraging the robust 6D pose estimation of objects and robot proprioception.
Specifically, we first introduce the interaction-aware point clouds, which establish correspondences between the object and the hand. These point clouds are then used for our pre-training policy, where we also incorporate object-centric contact maps and hand-arm coordination information, effectively capturing both spatial and temporal dynamics. 
Our method demonstrates exceptional dexterous manipulation capabilities, achieving state-of-the-art performance in six real-world tasks, surpassing other baselines by a large margin. Experimental results also highlight the superior generalization and robustness of CordViP to different objects, viewpoints, and scenarios. Code and videos are available on \textcolor{purple}{\url{https://aureleopku.github.io/CordViP}}.
\end{abstract}

\IEEEpeerreviewmaketitle

\section{Introduction}
Dexterous manipulation
is a fundamental capability in human daily life such as assembling small parts and opening boxes. Achieving human-level dexterity in real-world scenarios is crucial for integrating robots into everyday human activities. Recent advancements in imitation learning have demonstrated significant potential in various robotic manipulation tasks.  Some existing methods leverage 2D images as input to directly predict actions \cite{chi2023diffusion, zhao2023learning, wang2023mimicplay, xia2024cage}. While these vision-based imitation learning approaches are capable of handling a wide range of tasks, they typically demand extensive demonstrations \cite{haldar2023teach, chi2023diffusion, gervet2023act3d, wang2024equivariant} and, at the same time, 
fail to capture intricate spatial relationships and 3D structures essential for dexterous manipulation \cite{ze20243d, jia2024lift3dfoundationpolicylifting, shridhar2023perceiver, eisner2022flowbot3d}, limiting their ability to perform complex, fine-grained tasks.

Recent research has increasingly focused on 3D imitation learning, utilizing point clouds and voxels as representations, to help robots perceive 3D  environments and reason about spatial relationships \cite{ze20243d, gervet2023act3d, wang2024rise, goyal2023rvt, xian2023chaineddiffuser, su2024motion}. For example, 3D diffusion-based policies, which aim to enhance robots' ability to perform complex manipulation tasks by providing a more accurate and holistic representation of the environment.
These methods for obtaining 3D representations in real-world scenarios often depend on a single-view RGB-D camera to generate point clouds, imposing significant demands on both the camera's quality and its position.
However, the multi-fingered nature of dexterous hands often leads to occlusions during object manipulation, which not only loses the spatial information of the object point clouds but also obscures critical correspondence information for precise manipulation.
Besides, some works have explored the use of tactile sensors to enhance contact information \cite{guzey2023dexterity, yin2023rotating, guzey2024see, wu2024canonical, bhirangi2023all}. For instance, \citet{wu2024canonical} proposed a canonical representation of 3D tactile data, which is concatenated with other visual information.
Although tactile sensors hold promise, their high cost and susceptibility to interference from factors such as temperature fluctuations and electromagnetic fields undermine their stability and limit their practical applicability in real-world scenarios.
Amidst all the challenges, leveraging efficient and robust 3D imitation learning to understand spatial information in complex, dynamic environments is crucial for dexterous manipulation.

In this paper, we propose a \textbf{correspondence-based visuomotor policy (CordViP)}, which focuses on the spatial and temporal consistency between the manipulated object and the dexterous hand, even under significant occlusions. CordViP separately extracts the 3D representations of robotic hand's joint structure and the manipulated object, and utilizes the contact map between them as well as arm-hand cooperative motion information to pretrain the observation encoder. This approach enhances the dexterous hand's ability to understand spatial interactions and better facilitate a range of downstream tasks. Specifically, our framework operates in three phases: 
(1) We leverage the robust 6D pose estimation of objects combined with the robot's proprioceptive state to construct interaction-aware point clouds, providing ideal 3D observations that facilitates effective learning and inference for the visuomotor policy.
(2) We pretrain the encoder network on the play data we collected to predict the contact map between the dexterous hand and the object, and reconstruct the cooperative relationship between the hand and the arm for further understanding of spatial and temporal correspondences. 
(3) We utilize the pre-trained encoder to extract the semantic features of 3D point clouds and robot state, which encapsulates spatial consistency, contact dynamics and collaborative information, and then are used as conditions for the diffusion policy to predict actions. 

To comprehensively evaluate our proposed CordViP, we conduct extensive real-world experiments on six dexterous tasks: PickPlace, FlipCup, Assembly, Artimanip, FlipCap and LongHoriManip. Comparative results demonstrate that our method not only exhibits superior effectiveness, but also achieves remarkable performance with a minor number of expert demonstrations, highlighting its capability to learn efficiently from limited data. Furthermore, we find that CordViP generalizes well to various environmental perturbations, including varying lighting conditions, unseen objects, diverse scenarios, and different camera viewpoints, outperforming other baselines by a large margin.

In summary, our contributions are as follows:

\begin{enumerate}
    \item We develop a pipeline based on robust 6D pose estimation of objects and robot proprioception. This pipeline enables the real-time acquisition of complete 3D representations and semantic flow in real-world environments, and can effectively address the challenges posed by occlusions during dexterous manipulation.
    \item We propose CordViP, a correspondence-based visuomotor policy that utilizes the contact map and hand-arm coordination information to facilitate the understanding of spatial and temporal consistency.
    \item We demonstrate the effectiveness and generalization of our method through a range of real-world experiments using a dexterous hand.
\end{enumerate}

\section{Related Work}

\subsection{Dexterous Manipulation}
Dexterous Manipulation is a long-standing research topic in robotics that aims to give robots the ability to perform delicate operations like humans \cite{chen2022towards, bai2014dexterous, qin2021dexmv, touch-dexterity}. 
Traditional methods often rely on trajectory optimization based on dynamic models to solve operational problems \cite{kumar2014real, mordatch2012contact, wang2022dexgraspnet}, but these methods have limitations in complex tasks because they simplify the contact dynamics and are difficult to deal with uncertainties in dynamic environments. 
In contrast, Reinforcement Learning (RL) does not rely on accurate physical models but learns operational policies through interaction with the environment, which is highly adaptable. 
RL has achieved remarkable results in many dexterous manipulation tasks, such as object reorientation \cite{chen2022system, qi2023hand, pitz2023dextrous, handa2023dextreme} and sequential manipulation \cite{chen2023sequential, gupta2021reset}.
However, RL methods often suffer from several challenges, such as the need for extensive reward engineering and system design, as well as limited generalization to unseen scenarios.
Additionally, while Sim-to-Real is a common technique employed in RL, the gap between simulations and the real world degrades the performance of the policies once the models are transferred into real robots \cite{zhao2020sim}.
Imitation learning (IL), as another effective learning method, can quickly learn effective control policies by imitating expert demonstrations \cite{10602544, wang2023mimicplay}. 
In this work, we propose a correspondence-based visual imitation learning policy that utilizes spatial information between various components, enabling the acquisition of complex skills with a minimal number of expert demonstrations. 

\subsection{Imitation Learning}
Imitation learning (IL) allows a robot to directly learn from experts. Behavioral Cloning (BC) is one of the simplest imitation learning algorithms, which treats the problem of learning behavior as a supervised learning task \cite{pomerleau1988alvinn}.
The modeling methods commonly used in traditional BC, such as MSE, discretization \cite{lin2020limitations}, and K-Means \cite{guss2021towards}, have limitations when modeling complex action distributions. They fail to effectively capture the diversity and nuances of human behavior \cite{pearce2023imitating}. 
Over the past few years, diffusion models have emerged as a new modeling approach in BC, becoming powerful tools that enable robots to learn from demonstrations, handle uncertainty, and perform complex multi-step tasks with precision.
From the early applications of DDPMs to the recent innovations in BESO \cite{reuss2023goal}, OCTO \cite{team2024octo}, and CrossFormers \cite{doshi2024scaling}, these models have continually pushed the boundaries of what's possible in robotic behavior generation.
While traditional BC policies typically rely on 2D image-based representations \cite{chi2023diffusion, zhao2023learning, wang2023mimicplay, xia2024cage, liang2024dexhanddiffinteractionawarediffusionplanning,liu2025avr}, recent advancements have extended imitation learning to 3D visual representations \cite{ze20243d, chen2024g3flow, gervet2023act3d, wang2024dexcap, wang2024rise, lu2024manicm}. 
These 3D approaches provide a more comprehensive understanding of spatial relationships and 3D structures, further enhancing robotic behavior learning.

\subsection{Correspondence Learning}
Correspondence refers to the relationship or alignment between different entities or components, with the aim of establishing meaningful connections. Correspondence learning has been shown to improve performance in various robotic tasks, including grasping \cite{patten2020dgcm, ding2024preafford}, perception \cite{lai2021functional, chen2024g3flow}, pose estimation \cite{haugaard2022surfemb} and garment manipulation \cite{Wu_2024_CVPR}. In this paper, correspondence specifically refers to the alignment between hand-object spatial interaction and hand-arm temporal coordination. By incorporating correspondence, we enhance feature extraction capabilities, thereby enabling more accurate and coordinated movements in downstream tasks.



\section{Method}

\begin{figure*}
    \centering
    \includegraphics[width=\linewidth]{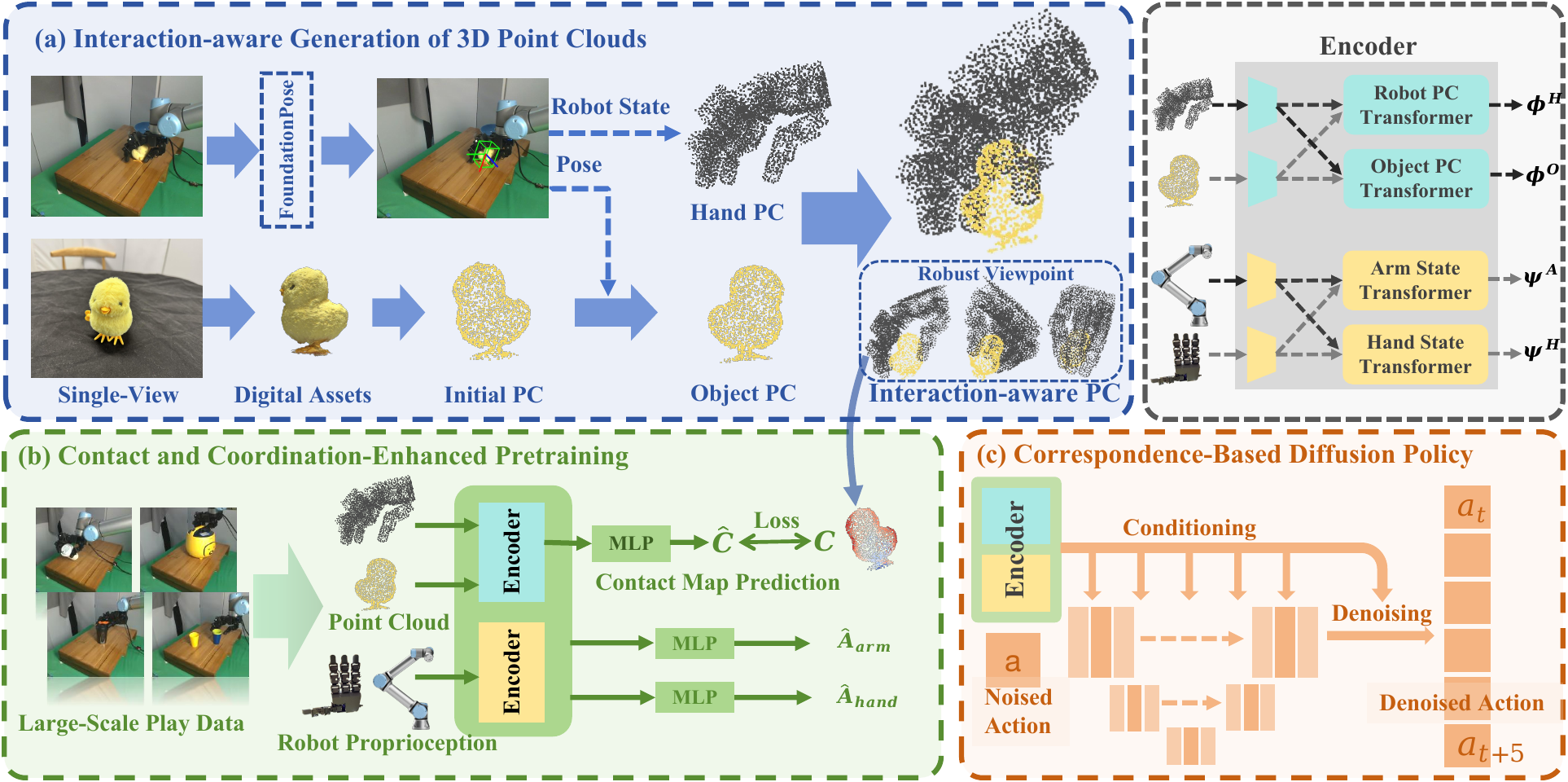}
    \caption{\textbf{Overview Framework}  (a) We first employ TripoSR to generate the initial object point cloud and FoundationPose to estimate the 6D pose of the object. In parallel, the hand point cloud is generated based on the robot's state. They are combined to construct interaction-aware point clouds, which demonstrate robustness to viewpoint variations. (b) During the pre-training phase, the generated point cloud data, combined with the robot’s proprioceptive information, is utilized to enhance spatial understanding and interaction modeling. (c) The pre-trained encoder is subsequently integrated into an imitation learning framework to facilitate downstream tasks in dexterous manipulation.}
    \label{fig:pipeline}
\end{figure*}

The overview of our framework is shown in Figure \ref{fig:pipeline}, which operates in three phases: (1) Interaction-aware generation of 3D point clouds. We acquire relatively accurate and complete 3D observations during real-world dexterous manipulation tasks even under significant occlusions, as described in \ref{sec:Interaction-aware Generation of 3D Point Clouds}.
(2) Contact and coordination-enhanced feature extraction. By leveraging large-scale play data and incorporating contact maps and hand-arm coordination, we improve spatial interaction perception and capture cooperative motion features, detailed in \ref{sec:Contact and Coordination-Enhanced Feature Extraction}. 
(3) Correspondence-based diffusion policy. The pre-trained encoder is used to extract 3D representations, which guide the training of a visuomotor policy, as outlined in \ref{sec:Correspondence-based Diffusion Policy}.

\subsection{Problem Formulation}
We formulate our problem as learning a visuomotor policy $\pi: \mathcal{O} \mapsto \mathcal{A}$ from expert demonstrations of the form of $\{(o_1, a_1), (o_2, a_2),\dots, (o_n, a_n)\}$, where $\mathcal{O}$ represents the robot's observations and $\mathcal{A}$ represents the corresponding actions, allowing the robot to generalize beyond the training data distribution. In our approach, each observation $o_t$ is composed of the object's point cloud $\mathbf{P}_{obj}$, the hand's point cloud $\mathbf{P}_{hand}$, and the robot's joint states, including a 6-Dof arm and 16-Dof hand configuration. 
Unlike previous works that rely on global point clouds for 3D feature extraction, our approach prioritizes capturing the individual information of the arm, hand, and object throughout the manipulation process. As a result, CordViP not only effectively addresses occlusion challenges during dexterous manipulation but also significantly improves the model's ability to comprehend spatial interactions and collaborative dynamics.
Furthermore, leveraging these observations, we compute contact maps between the robotic hand and the manipulated objects, as well as capture collaborative interaction information between the arm and hand. These elements are critical for modeling spatial and temporal relationships.

 
\subsection{Interaction-aware Generation of 3D Point Clouds}
\label{sec:Interaction-aware Generation of 3D Point Clouds}
Motivated by the superior generalization and efficiency of the 3D-based diffusion policy\cite{ze20243d, wang2024rise, ze2024generalizable, chen2024g3flow}, the key intuition behind our solution is to focus on the interactions between the hand and the manipulated object in 3D space. Although intuitively reasonable, achieving this goal is challenging in practice. On the one hand, real-world point cloud data, typically captured using stereo cameras or low-cost RGB-D scanners, suffers from geometric and semantic loss due to factors such as light reflection, material transparency, and limitations of sensor resolution and viewing angle. On the other hand, during dexterous manipulation with multi-fingered hands, occlusions frequently occur, resulting in the loss of critical contact and interaction information, which is vital for precise and effective manipulation. To this end, we propose the interaction-aware generation of 3D point clouds, enabling the reconstruction of crucial spatial information.

\textbf{Real-to-Sim for Digital Twin Generation.} To achieve the goal of obtaining a complete and accurate static point cloud of the manipulated object, we aim to reconstruct the digital twin from a single-view image~\cite{mu2024robotwin,mu2025robotwindualarmrobotbenchmark}. 
Referring to the approach TripoSR~\cite{TripoSR2024}, we implement a 3D generation technique,
which utilizes its strong priors and broad understanding of visual concepts in the 3D world to generate 3D digital assets.
To ensure the accuracy of point cloud flow tracking, we maintain consistency between the geometric and material properties of the reconstructed assets and their real-world counterparts. 
Subsequently, we uniformly sample points on the surface of the generated digital twin to obtain the initial 3D point clouds, providing a robust and accurate initial observation for both 3D spatial perception and pose tracking.

\textbf{Pose-Driven Point Cloud Tracking.}
We have successfully obtained the initial point cloud of the object. However, tracking the object’s point cloud in complex, real-world environments poses a significant challenge, particularly in scenarios with severe occlusions. To overcome this, we leverage foundation models to ensure precise pose estimation of the manipulated object, thereby enhancing point cloud tracking. Specifically, we first employ the Segment Anything model \cite{kirillov2023segany} to extract the mask of the manipulated object. This is combined with a digital twin generated via real-to-sim techniques, enabling us to use FoundationPose \cite{foundationposewen2024} for accurate pose estimation. Using the center of the acquired point cloud as the origin, we then apply the estimated pose to transform the entire point cloud, aligning it with the desired coordinate frame.

\textbf{Robot Point Cloud Forward Kinematics Model.}
To obtain the point cloud of the dexterous hand, we develop the robot point cloud forward kinematics model $\mathcal{F}_{pc}$. We first parse the URDF file to identify individual links of the robot system, and uniformly sample point clouds on the surface for each link. The robot system model is constructed with point clouds denoted as $\{P_{\ell_i}\}_{i=1}^{N_\ell}$, where $N_{\ell}$ is the number of links. Following \citet{wei2024dro}, we designed the robot point cloud forward kinematics model that maps any joint configuration to the corresponding pose of the point cloud. In order to focus on the interaction between the hand and the object, we ignore the point cloud of the robotic arm. Therefore, the 3D observation of the hand $\mathbf{P}_{hand} \in \mathbb{R}^{N_{\mathcal{P}} \times 3} $ is defined as:
\begin{equation}
    \mathbf{P}_{hand} = \mathcal{F}_{pc}(q, \{P_{\ell_i}\}_{i=1}^{N_\ell}),
\end{equation}
where $N_{\mathcal{P}}$ represents the downsampled size of the point cloud, which is set to 1024 in practice. Through our practical evaluations, the point cloud forward kinematics model operates at 8Hz, a frequency that is sufficiently high for our experimental setup, where inference is conducted at 5Hz.

\begin{figure}[t]
    \centering
    \includegraphics[width=\linewidth]{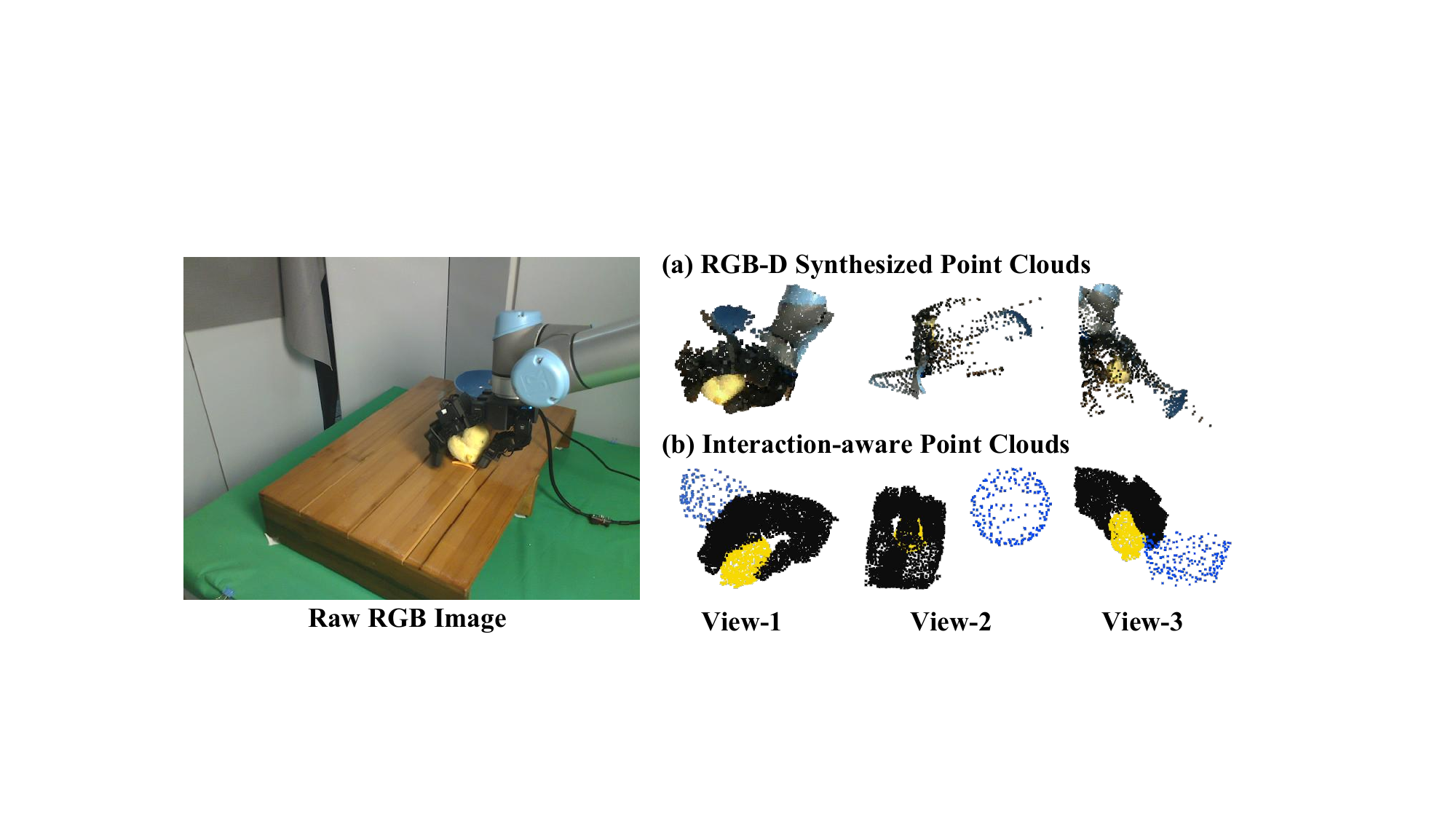}
    \caption{\textbf{Point Clouds Comparison.} We present point clouds of two methods under three different viewpoints. Notably, for better visualization, we have applied color information to the point clouds. However, color information is not used in the policy learning.}
    \label{fig:pointcloud compare}
\end{figure}

As shown in Figure \ref{fig:pointcloud compare}, interaction-aware point clouds significantly enhance the quality of 3D observations compared to RGB-D synthesized point clouds, while demonstrating strong robustness to different viewpoints.

\subsection{Contact and Coordination-Enhanced Feature Extraction}
\label{sec:Contact and Coordination-Enhanced Feature Extraction}
Interaction-aware Generation of 3D Point Clouds provides us with accurate and complete point cloud observation. However,  fine-grained dexterous manipulation often requires more detailed information. pre-training based on contact and coordination, in the context, can encourage the encoder to learn the intrinsic structures, facilitating more effective feature extraction, and promoting the learning of various downstream tasks. 

\textbf{Contact Map Synthesis.} A contact map serves as a critical piece of information in dexterous manipulation tasks, which captures the interaction between the hand and the object. Given the complete point clouds, we first calculate object-centric contact map $\mathcal{C}$ as the normalized distances from the object's surface point to the hand surface. Given the point clouds of the object $\mathbf{P}_{obj}$ and the point cloud of the hand $\mathbf{P}_{hand}$, the aligned distance $\mathcal{D_{(O, H)}}$ between each point $v_o$ on the object surface and the surface of the dexterous hand is defined as follows:
\begin{equation}
    \mathcal{D_{(O, H)}} = \min_{v_h \in \mathbf{P}_{hand}} e^{\gamma(1 - |<v_{h} - v_{o}, n_o>|)} \|v_o - v_h\|_2,
\end{equation}
where $n_o$ is the surface normal of the object, which is computed using the K-Nearest Neighbors methods \cite{1053964} by considering the local geometric properties of the point cloud. $\gamma$ is the scaling factor, which is empirically set to 1.

Based on the aligned distance, we compute the contact map following \citet{jiang2021graspTTA}:
\begin{equation}
    \mathcal{C} = 1 - 2(\text{Sigmod}(\theta \cdot \mathcal{{D_{(O, H)}}})-0.5)
\end{equation}
where $\theta$ is the scaling factor and each point's contact value $c_i \in \mathcal{C}$ is bounded within the [0, 1] range.

\textbf{Point Cloud Feature Extraction.}
We employ two encoders with identical architecture to extract point cloud embeddings, denoted as $f_{\theta_O}(\mathbf{P}_{obj})$ and $f_{\theta_H}(\mathbf{P}_{hand})$, for the object and hand, respectively. In detail, we adopt PointNet \cite{qi2016pointnet} as the point cloud encoder, which excels in capturing local structures and integrating global information. To establish correspondences between the hand and the object features, we apply two multi-head cross-attention transformers $g_{\theta_O}$, $g_{\theta_H}$ to fuse their respective embeddings, which maps the hand and object features to two sets of aligned representations, denoted as $\phi^R$ and $\phi^O$:
\begin{equation}
\begin{split}
    \phi^H &= g_{\theta_H}(f_{\theta_H}(\mathbf{P}_{hand}), f_{\theta_O}(\mathbf{P}_{obj})) + f_{\theta_H}(\mathbf{P}_{hand})  \\
    \phi^O &= g_{\theta_O}(f_{\theta_O}(\mathbf{P}_{obj}), f_{\theta_H}(\mathbf{P}_{hand})) + f_{\theta_O}(\mathbf{P}_{obj})
\end{split}
\end{equation}

To help the point cloud encoder learn the intrinsic features, we designed a contact map prediction task. Since 3D point cloud observations implicitly contain contact information, we utilize a three-layer MLP to predict the object-centric contact map $\mathcal{C}^{pred}$ given the point cloud observations of both the hand and the object. The MSE loss is calculated between $\mathcal{C}$ and $\mathcal{C}^{pred}$. This pre-training approach enables the encoder to learn the interactions and relationships within the environment.

\textbf{Hand-Arm Coordination Enhancement.}
To help the robot system learn the features of hand-arm coordination, we also propose a correspondence-based design for action prediction. The arm and hand states are first projected into vectors of identical dimensionality through a linear layer, after which the same cross-attention transformers are employed to establish correspondences between the hand and the arm. We predict the action sequence of the robot arm based on the point clouds and the state of the hand. Similarly, we also predict the action sequence of the hand using point clouds and the arm state. We use MSE loss to compute the loss between the reconstructed and original action. By further predicting the action sequence respectively, the encoder is able to learn intrinsic features of motion and capture collaborative dynamics.

Given the aforementioned losses, our overall training objective during the pre-training phase is defined as:
\begin{equation}
    \underset{\mathcal{E}}{\min} \mathcal{L} = \mathcal{L}_{contact} + \lambda \mathcal{L}_{coordination}, 
\end{equation}
where $\mathcal{E}$ represents the encoder of the observation, and $\lambda$ is a hyperparameter that controls the relative strengths of the losses.

\subsection{Correspondence-based Diffusion Policy}
\label{sec:Correspondence-based Diffusion Policy}

After obtaining the pre-trained encoder, we utilize an imitation learning framework to learn visuomotor policy for dexterous manipulation tasks. Specifically, we adopt conditional denoising diffusion model \cite{ho2020denoising, chi2023diffusion, pearce2023imitating} as our backbone, which conditions on 3D visual features $\phi^{H,O}$ and robot states features $\psi^{A,H}$. Beginning with a Gaussian noise $A^K$, the denoising network $\epsilon_\theta$ performs $k$ iterations to gradually denoise $A^K$ into the noise-free action $A^0$:
\begin{equation}
        A^{k-1} = \alpha_k(A^k - \gamma_k \varepsilon_{\theta}(\phi^{H,O}, \psi^{A,H}, A^k, k)) + \sigma_k \mathcal{N}(0,I),
\end{equation}
where $\mathcal{N}(0,I)$ is Gaussian noise, $\alpha_k$, $\gamma_k$ and $\sigma_k$ are functions of $k$, determined by the noise scheduler. This formulation allows the model to capture the distribution of action without the cost of inferring future states.

We use the DDIM scheduler \cite{song2020denoising} to accelerate the inference speed in real-world experiments. The training objective is to predict the noise added to the original data:
\begin{equation}
    \mathcal{L} = MSE(\varepsilon^k, \varepsilon_{\theta}(\bar{\alpha_k}A^0+\bar{\beta}\varepsilon^k, \phi^{H,O}, \psi^{A,H}, k))
\end{equation}

Unlike the original diffusion-based policy, we incorporate consistency-related features as a condition for policy learning using the pre-trained encoder and fine-tuning the encoder during downstream tasks training. This encoder implicitly extracts contact and coordination information, thereby enhancing the policy's understanding of spatial relationships.

\section{Experiments}
 We conduct comprehensive real-world experiments to answer the following questions:
 \begin{itemize}
     \item To what extent can our framework promote the learning of the visuomotor policy for dexterous manipulation across diverse real-world scenarios (Section \ref{sec: Effectiveness})? 
     \item How promising is CordViP in terms of sample efficiency and generalization capability (Section \ref{sec: Efficiency}, \ref{sec: Generalization})?
     \item What role does each of the system components play in enhancing its overall performance (Section \ref{sec: Ablation}, \ref{sec: Transferability})?
 \end{itemize}

\subsection{Experiment Setup}
\label{sec: Experiment Setup}
\begin{figure}
    \centering
    \includegraphics[width=\linewidth]{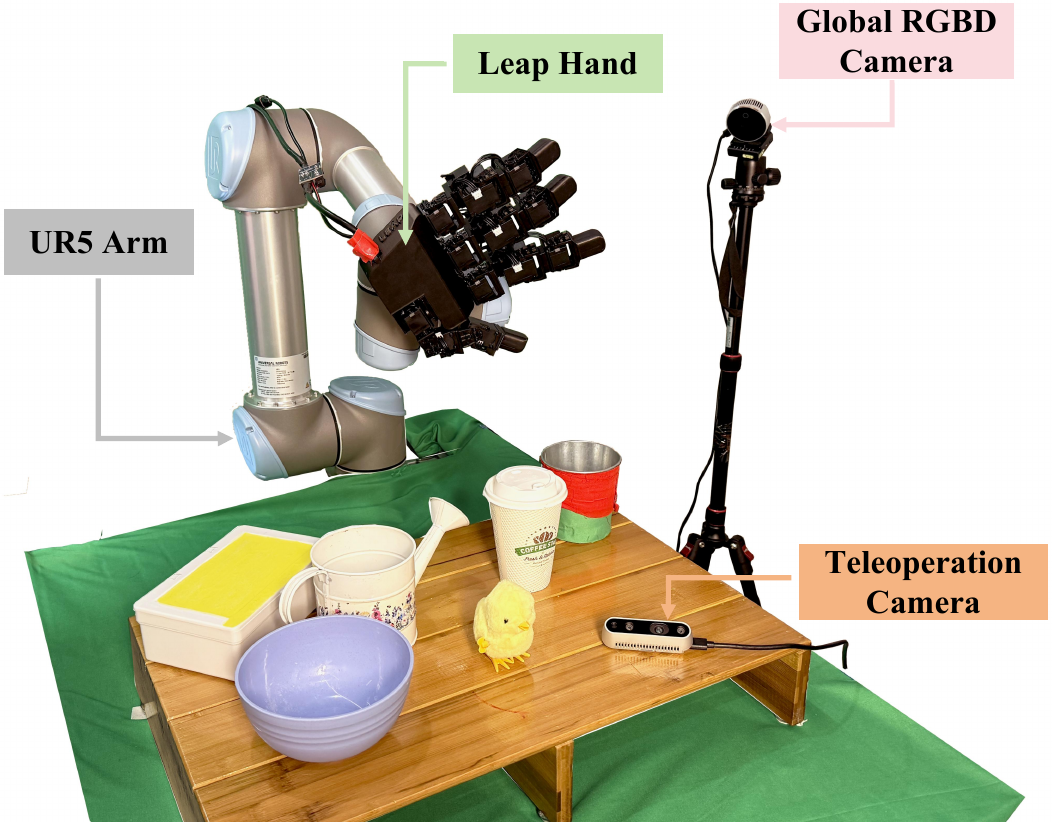}
    \caption{\textbf{Real robot system.} Our system consists of a Leap Hand and a UR5 Arm, with a fixed Realsense L515 camera employed to capture visual observation. The Realsense D435 camera is only used for data collection during teleoperation and is not involved in the policy learning.}
    \label{fig:robot system}
\end{figure}

\textbf{Robot System Setup.} As shown in Figure \ref{fig:robot system}, our system consists of a 6-Dof UR5 robot arm and a 16-Dof Leap Hand \cite{shaw2023leap} with four fingers. A single Intel Realsense L515 RGBD camera is mounted on the side of the robot to capture visual observation.

\begin{figure*}[t]
    \centering
    \includegraphics[width=1\linewidth]{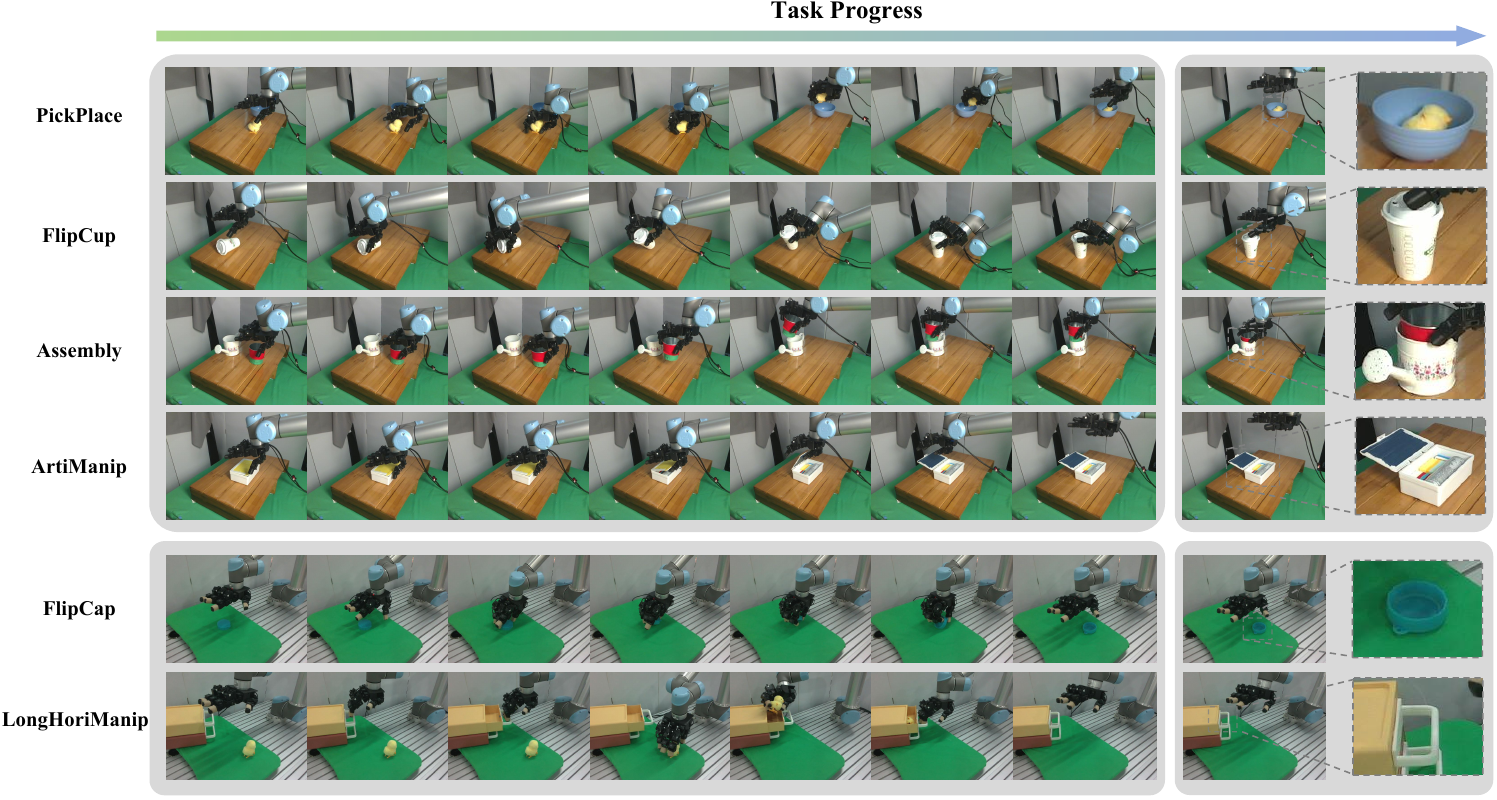}
    \caption{\textbf{Visualization of six dexterous manipulation tasks}, with the right side showing the end state.}
    \label{fig:task_episode}
\end{figure*}

\textbf{Tasks.}
We evaluate our approach on four base dexterous manipulation tasks, along with two advanced contact-rich and fine-grained tasks, as shown in Figure \ref{fig:task_episode}. The episode length of each task will be limited to a maximum of 500 steps and each task is evaluated with 20 trials by default. We now briefly describe our tasks:
\begin{enumerate}
    \item \textbf{PickPlace.} The Leaphand picks up a toy chicken and places it into a blue bowl.
    \item \textbf{FlipCup.} The Leaphand reaches a cup lying on the table, lifts it up, and then rotates the wrist to position the cup upright on the table.
    \item \textbf{Assembly.} The Leaphand reaches and grasps a cylindrical cup, and insert it into a kettle.
    \item \textbf{ArtiManip.} The Leaphand lifts the lid of a box using its thumb and gently opens it, which involves the manipulation of the articulated objects.
    \item \textbf{FlipCap.} This task requires four-finger coordination: the thumb and middle finger lift and rotate the cap slightly, while the index finger pushes it from the opposite side to complete the flip.
    \item \textbf{LongHoriManip.} This task involves four sequential steps: pull, pick, place, and push, requiring precise and continuous control across a long horizon.
\end{enumerate}

\begin{figure}[t]
    \centering
    \includegraphics[width=0.95\linewidth]{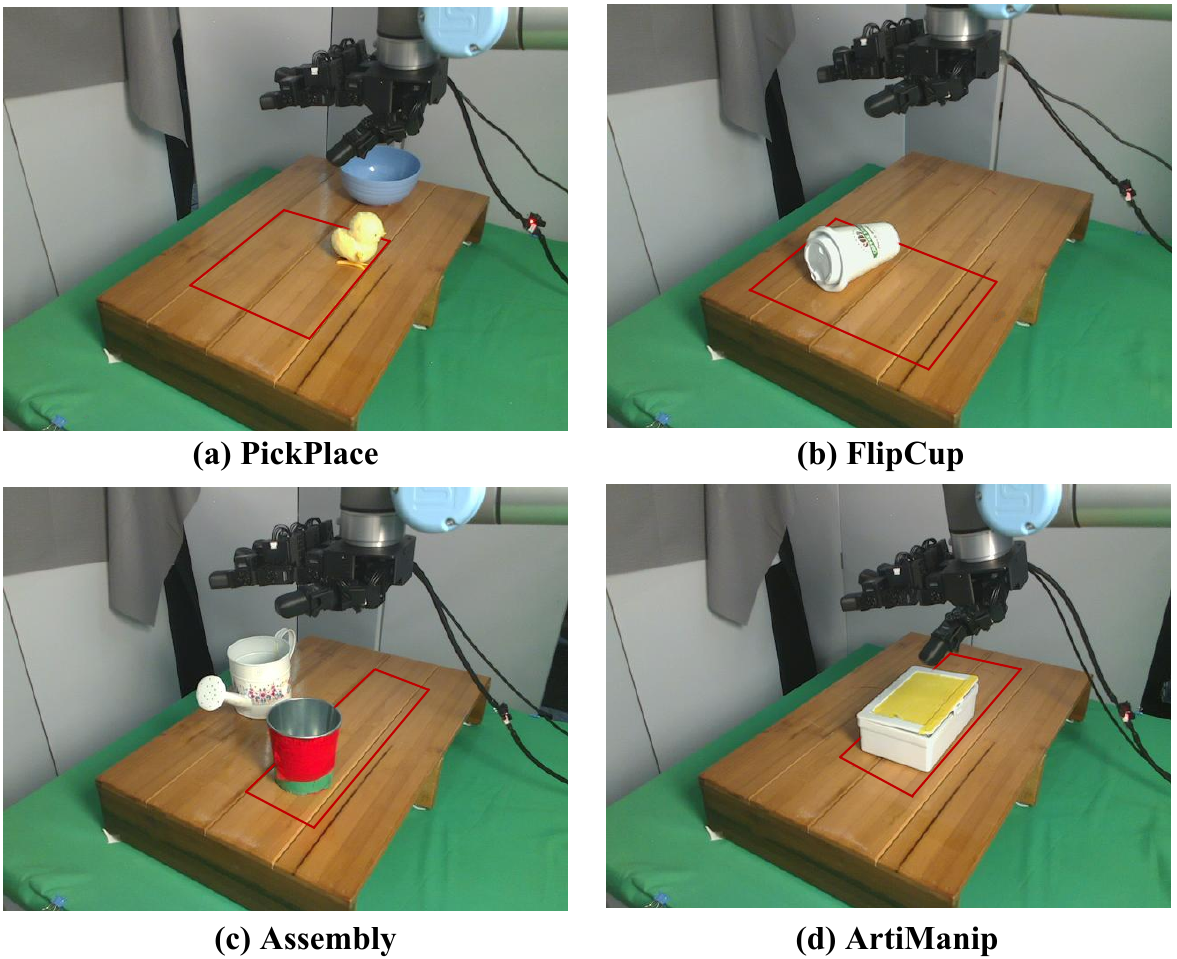}
    \caption{\textbf{Randomization of Object Positions.} The red rectangles mark the range of positions of manipulated objects. For PickPlace and FlipCup, both the toy chicken and the cup are randomly rotated within a certain range.}
    \label{fig:randomization}
\end{figure}

For each task, we incorporate a certain level of randomization to ensure that the policy learns the task-specific features rather than fitting the trajectory. As shown in Figure \ref{fig:randomization}, the objects are randomly placed within the red rectangular region. More details are provided in Appendix \ref{appendix: Real-World Task Description}.

\textbf{Expert demonstrations.} Our training demonstrations are collected by human teleoperation. Since our tasks place significant emphasis on coordination and coherence between the hand and the arm, we employ a vision-based approach to teleoperate the robot. Specifically, we use HaMeR \cite{pavlakos2024reconstructing} to track human hand pose with a single Realsense D435 camera and use Anyteleop \cite{qin2023anyteleop} framework to retarget the robot system. The robotic arm is controlled through the 6-Dof end-effector's pose while the robotic hand is controlled through the retargeted hand joint positions. We collect 50 demonstrations for each task, with all expert demonstrations recorded at a frequency of 5Hz.

\textbf{Baselines.} We first compare our method with six state-of-the-art imitation learning algorithms, \ie three vision-based policies(BCRNN \cite{mandlekar2021matters}, ACT \cite{zhao2023learning}, DP \cite{chi2023diffusion}) and three 3D-based policies(BCRNN+3D, ACT+3D, DP3 \cite{ze20243d}). BCRNN+3D and ACT+3D are variants of BCRNN and ACT, respectively, in which the image input is substituted with point clouds, encoded using PointNet.We emphasize that the resolution of both the image and depth for all 2D and 3D baseline methods is kept consistent, and the point clouds are synthesized from the RGBD camera.

To further assess the performance under the same privileged setting as CordViP, we also introduce three additional baselines: (1) State-based MLP, a simple behavior cloning policy that takes the robot's proprioception and the object pose as input; (2) State-based Diffusion Policy, which conditions a diffusion model on the concatenated proprioceptive and object pose features; and (3) G3Flow \cite{chen2024g3flow}, a recent method that dynamically computes semantic flow based on pose information and integrates it with raw 3D point clouds in the Diffusion Policy framework.

\subsection{Effectiveness}
\label{sec: Effectiveness}
The results of the effectiveness experiments are given in Table \ref{tab:Effectiveness base} and Table \ref{tab:Effectiveness advanced}. Our proposed CordViP maintains a completion rate of over 85\% across four base tasks, significantly outperforming the other baselines, and also demonstrates superior performance on the two advanced tasks. BCRNN and its 3D variant perform poorly in all base tasks. While the dexterous hand can locate and reach the object, a substantial finger jitter is observed during the grasping and contact-rich manipulation phases. The image-based policies, ACT and DP, achieve good performance in the flipcup and pickplace tasks. However, they struggle with the assembly task. This is mainly due to the significant occlusions during manipulation, where these image-based policies fail to effectively leverage spatial information. state-based DP matches and even outperforms vision-based baselines on simpler tasks. However, due to the lack of geometric information, it struggles with more dexterous tasks. Compared to ACT, ACT+3D shows superior results in all tasks, highlighting the crucial importance of geometric structure and spatial information in policy learning. G3Flow, which benefits from both 3D visual and state information, is the strongest baseline overall. However, its performance on tasks such as FlipCap is limited due to the impact of 3D noise and the lack of correspondence modeling.

Interestingly, we discover that the recently proposed DP3 underperforms relative to DP in most of our tasks, which aligns with the findings of \citet{wang2024rise}. Although several studies have demonstrated the effectiveness of DP3 in simulation and relatively ideal real-world scenarios, such as those with multiple viewpoints, we observe that the quality of the point clouds has a significant impact on DP3's performance. Further details will be discussed in Section \ref{sec: Ablation}. In contrast, our method exhibits remarkable robustness to the quality of point clouds. By establishing correspondences between the dexterous hand and objects, CordViP facilitates more effective perception of spatial and interaction information.

\textbf{Inference Efficiency.} CordViP also attains an efficient inference speed, as it eliminates the need for the farthest point sampling of point clouds while utilizing compact 3D representations. We evaluate the inference speed of CordViP compared with DP3 on an Intel i7-14700KF CPU and RTX 4090D GPU, CordViP reaches a maximum of 12.84 FPS, surpassing DP3's 11.79 FPS. This highlights that our approach not only achieves enhanced performance but also maintains low computational overhead during inference.

\begin{table}
    \renewcommand{\arraystretch}{1.1} 
    \centering
    \caption{\textbf{Main results of four base real-world tasks.} Each experiment is evaluated with 20 trials.}
    \label{tab:Effectiveness base}
    \footnotesize
    \resizebox{\linewidth}{!}{
        \begin{tabular}{c|cccc|c}
            \toprule
            \bf{Method} & \bf{PickPlace} & \bf{FlipCup} & \bf{Assembly} & \bf{ArtiManip} & \bf{Avg}\\
            \midrule
            {State-based MLP}
            &  {0\%} & {0\%} & {0\%} & {0\%} & {0\%} \\
            {BCRNN}
            &  {0\%} & {0\%} & {0\%} & {5\%} & {1\%} \\
            {BCRNN+3D}
            & {0\%} & {5\%} & {0\%} & {10\%} & {4\%} \\
            {ACT}
            & {45\%} & {70\%} & {25\%} & {65\%} & {51\%} \\
            {ACT+3D}
            & {70\%} & {80\%} & {35\%} & {70\%} & {64\%} \\
            {DP}
            & {55\%} & {65\%} & {20\%} & {35\%} & {44\%} \\
            {DP3}
            & {30\%} & {20\%} & {0\%} & {40\%} & {23\%} \\
            {State-based DP}
            &  {40\%} & {35\%} & {40\%} & {40\%} & {39\%} \\
            {G3Flow}
            &  {65\%} & {65\%} & {80\%} & {85\%} & {74\%} \\
            \rowcolor{gray!20}
            {\bf{CordViP(Ours)}}
            & {\textbf{85\%}} & {\textbf{90\%}} & {\textbf{90\%}} & {\textbf{95\%}} & {\textbf{90\%}} \\
            \bottomrule
        \end{tabular}
    }
    \vspace{-1em}
\end{table}

\begin{table}
    \renewcommand{\arraystretch}{1.1} 
    \centering
    \caption{\textbf{Main results of advanced tasks.} Each experiment is evaluated with 10 trials.}
    \label{tab:Effectiveness advanced}
    \footnotesize
    \resizebox{\linewidth}{!}{
        \begin{tabular}{c|c|cccc}
            \toprule
            \multirow{2}{*}{\textbf{Method}} & \multirow{2}{*}{\textbf{FlipCap}} & \multicolumn{4}{c}{\textbf{LongHoriManip}} \\
            \cline{3-6}
             &  & \textbf{Pull} & \textbf{Pick} & \textbf{Place} & \textbf{Push} \\
            \midrule
            {ACT}
            & {30\%} & {\textbf{100\%}} & {40\%} & {40\%} & {30\%} \\
            {ACT+3D}
            & {40\%} & {\textbf{100\%}} & {40\%} & {20\%} & {20\%} \\
            {DP}
            & {30\%} & {70\%} & {30\%} & {10\%} & {10\%} \\
            {State-based DP}
            &  {20\%} & {80\%} & {30\%} & {10\%} & {0\%} \\
            {G3Flow}
            &  {60\%} & {\textbf{100\%}} & {80\%} & {50\%} & {30\%} \\
            \rowcolor{gray!20}
            {\bf{CordViP(Ours)}}
            & {\textbf{80\%}} & {\textbf{100\%}} & {\textbf{100\%}} & {\textbf{70\%}} & {\textbf{60\%}} \\
            \bottomrule
        \end{tabular}
    }
\end{table}

\subsection{Efficiency}
\label{sec: Efficiency} 
\begin{figure}
    \centering
    \includegraphics[width=\linewidth]{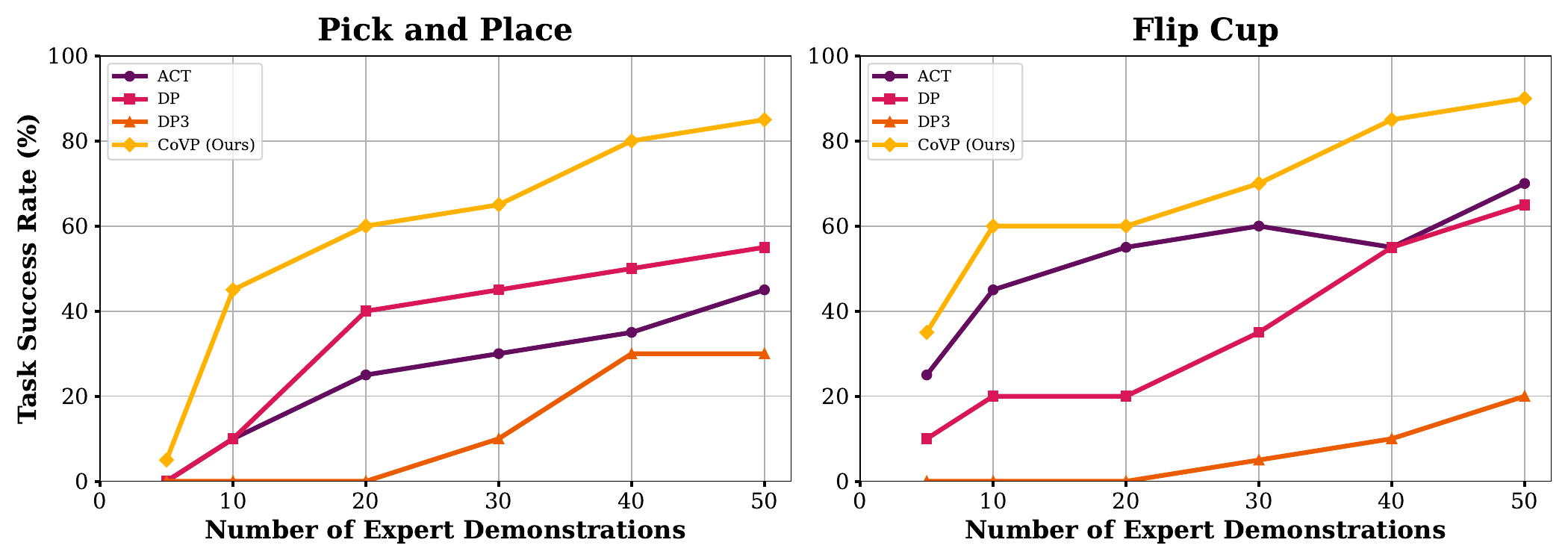}
    \caption{\textbf{Experimental results of efficiency.} We train ACT, DP, DP3, and CordViP on the PickPlace and FlipCup tasks with an increasing number of demonstrations.}
    \label{fig:efficiency}
\end{figure}

The number of expert demonstrations plays a crucial role in the performance of imitation learning. To access the learning efficiency, we train ACT, DP, DP3 and our proposed CordViP on dexterous tasks using varying quantities of expert demonstrations. As illustrated in Figure \ref{fig:efficiency}, CordViP exhibits superior performance, achieving higher accuracy with fewer demonstrations. Remarkably, even with just 10 demonstrations, CordViP effectively establishes correspondences, extracts spatial and geometric features, and maintains a high success rate. Furthermore, when provided with a sufficient number of demonstrations, CordViP also demonstrates excellent manipulation capabilities.

\subsection{Generalization}
\label{sec: Generalization}
Besides the remarkable effectiveness and efficiency, CordViP also showcases excellent generalization capabilities in real-world dexterous manipulation tasks. In this section, we comprehensively investigate its generalization abilities across four aspects, as detailed below.

\textbf{Generalize to different lighting conditions.}

We established three distinct lighting conditions. As shown in Table \ref{tab:generalization:lighting conditions}, the 3D representation of our proposed CordViP is dynamically maintained through FoundationPose tracking. FoundationPose, which integrates multimodal information (RGB images, depth data, and prior 3D model knowledge of the target object), demonstrated significant robustness to lighting variations in our empirical tests. In contrast, diffusion policies failed to complete tasks under most conditions. These image-based policies typically rely on data augmentation to improve generalization, which may introduce instability during training.

\begin{table}
    \renewcommand{\arraystretch}{1.1} 
    \centering
    \caption{\textbf{Generalization results on different lighting conditions.} We evaluate the policy under three lighting scenarios: dim light(dim), white light(white) and colored lighting(colored).}
    \vspace{-2em}
    \begin{figure}[H]
        \centering
        \includegraphics[width=\linewidth]{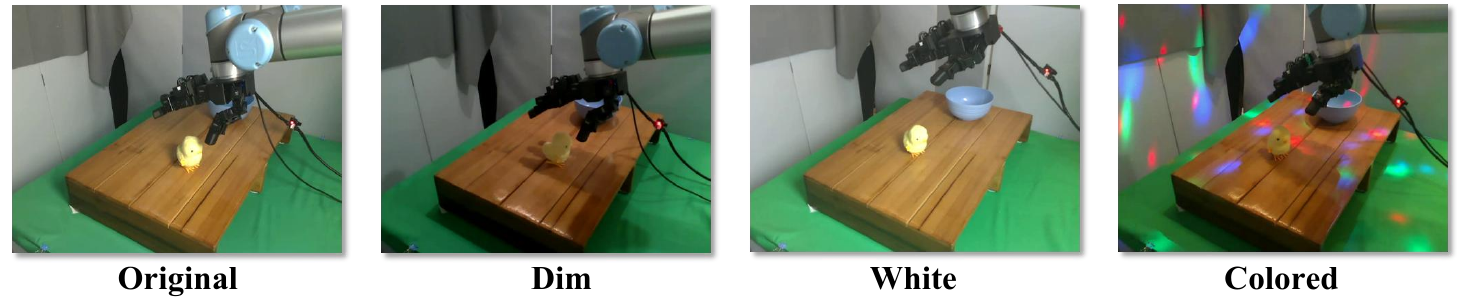}
        \label{fig:light}
    \end{figure}
    \vspace{-2em}
    \label{tab:generalization:lighting conditions}
    \footnotesize
    \resizebox{\linewidth}{!}{
        \begin{tabular}{c|ccc|ccc}
            \toprule
            \multirow{2}{*}{\bf{Method}} &  \multicolumn{3}{c|}{\bf{PickPlace}} & \multicolumn{3}{c}{\bf{FlipCup}} \\
            & Dim & White & Colored & Dim & White & Colored \\
            \midrule
            {DP}
            &  {0\%} & {25\%} & {0\%} & {0\%} & {50\%} & {0\%} \\
            {ACT+3D}
            & {65\%} & {55\%} & {70\%} & {75\%} & {80\%} & {75\%} \\
            {\bf{CordViP(Ours)}}
            & \textbf{80\%} & \textbf{85\%} & \textbf{80\%} & \textbf{80\%} & \textbf{85\%} & \textbf{90\%} \\
            \bottomrule
        \end{tabular}
    }
    \vspace{-1em}
\end{table}

\textbf{Generalize to different scenarios.}
CordViP relies on a robust 6D pose estimator to track objects and build interaction-aware 3D point clouds, which enables our method to generalize to different scenarios. The results presented in Table \ref{tab:generalization：scenarios} show that DP is highly sensitive to visual variations, while ACT+3D performs well in scenarios with different visual appearances due to its reliance on 3D geometric inputs. However, we observe 3D-based policy struggles in more challenging, cluttered scenes, as randomly placed objects can lead to noisy or incomplete point cloud representations.  In contrast, our method effectively focuses on the manipulation of subjects and objects, establishing spatial correspondences, and has demonstrated strong generalization to different scenarios.
\begin{table}[t]
    \renewcommand{\arraystretch}{1.1} 
    \vspace{2em}
    \centering
    \caption{\textbf{Generalization results in diverse scenarios}, including varying visual appearances and challenging cluttered environments.}
    \vspace{-2em}
    \begin{figure}[H]
        \centering
        \includegraphics[width=\linewidth]{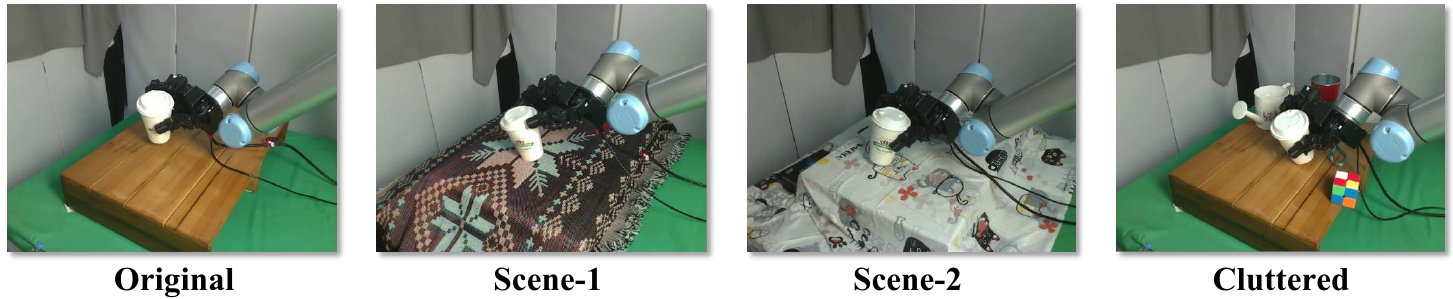}
        \label{fig:scene}
    \end{figure}
    \vspace{-2em}
    \label{tab:generalization：scenarios}
    \footnotesize
    \resizebox{\linewidth}{!}{
        \begin{tabular}{c|ccc|ccc}
            \toprule
            \multirow{2}{*}{\bf{Method}} &  \multicolumn{3}{c|}{\bf{PickPlace}} & \multicolumn{3}{c}{\bf{FlipCup}} \\
            & Scene-1 & Scene-2 & Cluttered & Scene-1 & Scene-2 & Cluttered \\
            \midrule
            {DP}
            &  {0\%} & {0\%} & {0\%} & {0\%} & {0\%} & {5\%} \\
            {ACT+3D}
            & {50\%} & {70\%} & {5\%} & \textbf{80\%} & {70\%} & {20\%} \\
            {\bf{CordViP(Ours)}}
            & \textbf{70\%} & \textbf{80\%} & \textbf{75\%} & {75\%} & \textbf{80\%} & \textbf{85\%} \\
            \bottomrule
        \end{tabular}
    }
    \vspace{-1em}
\end{table}

\textbf{Generalize to unseen objects.}
To validate policies' generalization to different objects, we test them on the PickPlace and FlipCup tasks using three unseen objects, each varying in color, shape, and dynamics. As shown in Table \ref{tab:generalization： unseen objects}, DP shows poor generalization to unseen objects, while ACT+3D shows a certain level of generalization ability. However, ACT+3D struggles to adjust effectively and promptly when confronted with objects that exhibit significant differences. CordViP demonstrates strong generalization ability when handling unseen objects, making fine-grained adjustments by establishing spatial and temporal correspondences. 

\begin{table}
    \renewcommand{\arraystretch}{1.1} 
    \centering
    \caption{\textbf{Generalization results on unseen objects.}
    For PickPlace and FlipCup tasks, we chose three previously unseen objects, varying in color, shape, and dynamics.}
    \vspace{-2em}
    \begin{figure}[H]
        \centering
        \includegraphics[width=\linewidth]{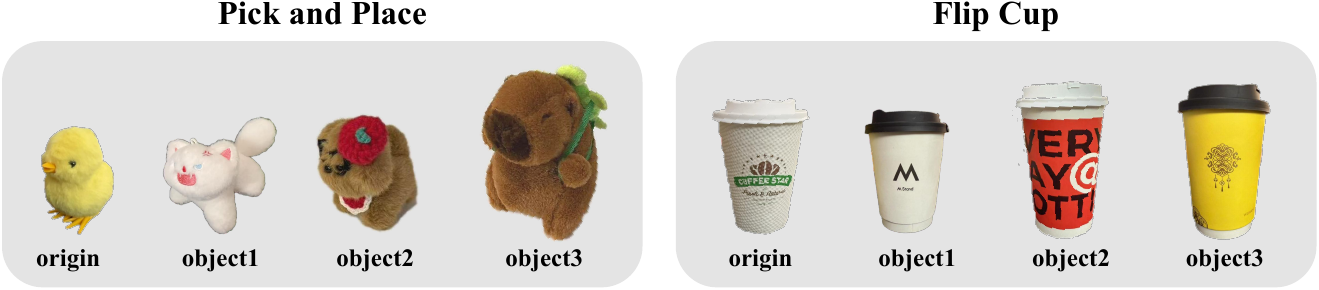}
        \label{fig:unseen_obj}
    \end{figure}
    \vspace{-2em}
    \label{tab:generalization： unseen objects}
    \footnotesize
    \resizebox{\linewidth}{!}{
        \begin{tabular}{c|ccc|ccc}
            \toprule
            \multirow{2}{*}{\bf{Method}} &  \multicolumn{3}{c|}{\bf{PickPlace}} & \multicolumn{3}{c}{\bf{FlipCup}} \\
            & Object-1 & Object-2 & Object-3 & Object-1 & Object-2 & Object-3 \\
            \midrule
            {DP}
            &  {0\%} & {5\%} & {0\%} & {15\%} & {35\%} & {0\%} \\
            {ACT+3D}
            & {55\%} & \textbf{85\%} & {75\%} & {55\%} & {40\%} & \textbf{75\%} \\
            {\bf{CordViP(Ours)}}
            & \textbf{75\%} & \textbf{85\%} & \textbf{90\%} & \textbf{60\%} & \textbf{45\%} & \textbf{75\%} \\
            \bottomrule
        \end{tabular}
    }
\end{table}

\textbf{Generalize to different viewpoints.}
Achieving generalization to different viewpoints presents a more significant challenge, as the expert demonstrations are all captured using a fixed camera. We evaluate DP, ACT+3D, and our method with three different camera viewpoints. For 3D-based methods, since the point clouds are represented in the world coordinate system, the camera is recalibrated for each viewpoint. As shown in Table \ref{tab:generalization：viewpoints}, the image-based diffusion policy is highly sensitive to camera viewpoints and completely fails across all three camera views. ACT+3D, which leverages 3D information, demonstrates a certain level of generalization to minor viewpoint changes. However, due to the limitations of the camera viewpoint, the synthesized point clouds can only capture partial spatial information, making it hard to handle significant changes in camera views. By leveraging comprehensive 3D priors of manipulated objects, our method achieves view-agnostic generation of interaction-aware point clouds with full spatial coverage, demonstrating robust performance under significant variations in camera perspectives.

\begin{table}
    \renewcommand{\arraystretch}{1.1} 
    \centering
    \caption{\textbf{Generalization results to different viewpoints.} }
    \vspace{-2em}
    \begin{figure}[H]
        \centering
        \includegraphics[width=\linewidth]{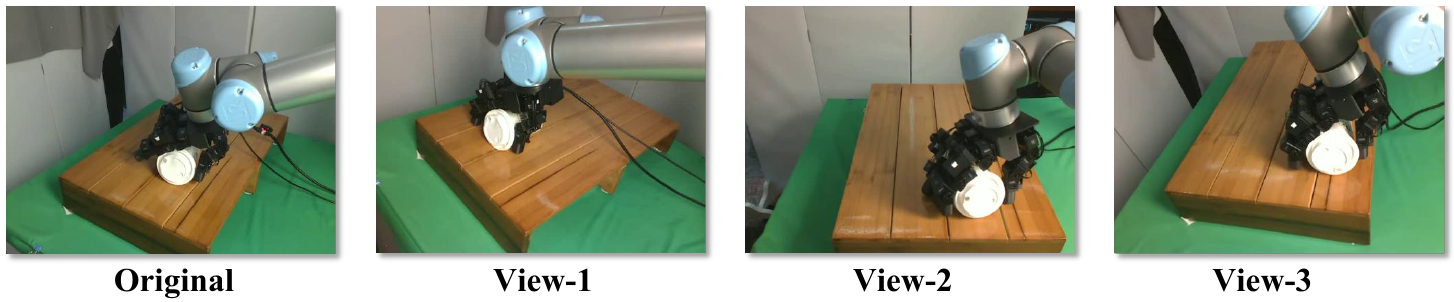}
        \label{fig:view}
    \end{figure}
    \vspace{-2em}
    \label{tab:generalization：viewpoints}
    \footnotesize
    \resizebox{\linewidth}{!}{
        \begin{tabular}{c|ccc|ccc}
            \toprule
            \multirow{2}{*}{\bf{Method}} &  \multicolumn{3}{c|}{\bf{PickPlace}} & \multicolumn{3}{c}{\bf{FlipCup}} \\
            & View-1 & View-2 & View-3 & View-1 & View-2 & View-3 \\
            \midrule
            {DP}
            &  {0\%} & {0\%} & {0\%} & {0\%} & {0\%} & {0\%} \\
            {ACT+3D}
            & {20\%} & {45\%} & {60\%} & {0\%} & {40\%} & {30\%} \\
            {\bf{CordViP(Ours)}}
            & \textbf{80\%} & \textbf{75\%} & \textbf{65\%} & \textbf{60\%} & \textbf{70\%} & \textbf{80\%} \\
            \bottomrule
        \end{tabular}
    }
    \vspace{-1em}
\end{table}

\begin{figure*}[h]
    \centering
    \begin{subfigure}{\textwidth}
        \captionsetup{skip=3pt}
        \centering
        \includegraphics[width=\textwidth]{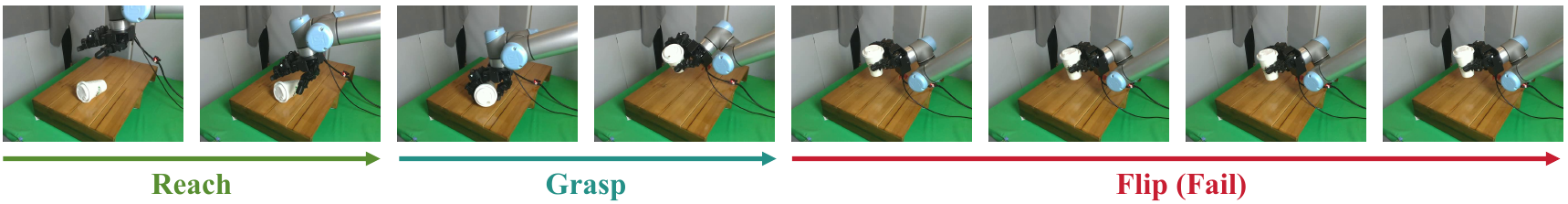}
        \caption{Case1: FlipCup.}
        \label{case1}
    \end{subfigure}
    \\
    \vspace{0.5em}
    \begin{subfigure}{\textwidth}
        \captionsetup{skip=3pt}
        \centering
        \includegraphics[width=\textwidth]{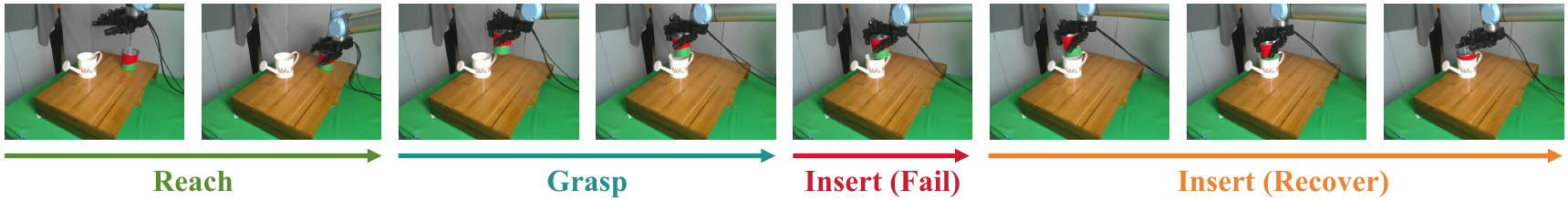}
        \caption{Case2: Assembly.}
        \label{case2}
    \end{subfigure}
    \vspace{-1em}
    \caption{\textbf{Failure case.} (a) Case 1 is a failure case from the Flip Cup task; (b) Case2 is an example from the Assembly task, where after a failure to insert the cup, the policy made adjustments, corrected the orientation of the cup, and successfully completed the task.}
    \label{fig: failure case}
\end{figure*}

\begin{table*}
    \renewcommand{\arraystretch}{1.1} 
    \centering
    \caption{\textbf{Ablation experiments on the Effectiveness of different components.} DP3 + Interaction-aware PC refers to using the Interaction-aware point clouds as the visual input for DP3. W/o. contact and coordination pretrain means that the encoder was not pre-trained with contact and coordination data.}
    \label{tab:ablation of components}
    \footnotesize
    \resizebox{0.85\linewidth}{!}{
        \begin{tabular}{c|cccc|c}
            \toprule
            \bf{Ablation} & \bf{PickPlace} & \bf{FlipCup} & \bf{Assembly} & \bf{ArtiManip} & \bf{Avg}\\
            \midrule
            {\textbf{DP3 + Interaction-aware PC}}
            &  {65\%} & {75\%} & {60\%} & {85\%} & {71\%} \\
            {\textbf{CordViP w/o. contact and coordination pretrain}}
            & {75\%} & {80\%} & {75\%} & {85\%} & {79\%} \\
            {\textbf{CordViP w/o. contact pretrain}}
            & {85\%} & {75\%} & {85\%} & {90\%} & {84\%} \\
            {\textbf{CordViP w/o. coordination pretrain}}
            & {80\%} & {85\%} & {80\%} & {85\%} & {83\%} \\
            {\textbf{CordViP(Ours)}}
            & {\textbf{85\%}} & {\textbf{90\%}} & {\textbf{90\%}} & {\textbf{95\%}} & {\textbf{90\%}} \\
            \bottomrule
        \end{tabular}
    }
\end{table*}

\subsection{Ablations}
\label{sec: Ablation}
\textbf{Effectiveness of components.} We conduct a series of ablation experiments on four base real-world tasks to evaluate the effectiveness of different components in our method. As shown in Table \ref{tab:ablation of components}, our results reveal the critical importance of contact and coordination information for learning dexterous manipulation policy. Specifically, for contact-rich tasks such as Flip-Cup, the absence of pre-training on contact information significantly reduces success rates. This highlights the essential role of contact in understanding physical interactions between the robotic hand and objects. On the other hand, for tasks that require continuous motion of both the arm and hand, such as Assembly and ArtiManip, incorporating coordination information notably enhances the policy's ability to execute complex movement patterns. These findings emphasize the necessity of both contact and coordination information to achieve robust manipulation.

We further observe that merging the Interaction-aware point clouds into a unified representation and providing it as the 3D input to DP3 leads to a noticeable improvement in the success rates across all tasks. This also suggests that, compared to ACT+3D, DP3 places higher demands on the quality of the point clouds, and is less robust to noise and occlusion in real-world scenarios. In our settings, we do not carefully position the camera to minimize occlusion of the dexterous hand, which demonstrates the robustness of our method to different viewpoints.

\textbf{Choice of point cloud encoders.} We use PointNet \cite{qi2016pointnet}, PointNet++ \cite{qi2017pointnetplusplus}, and PointNeXt \cite{qian2022pointnext} as point cloud encoders. As shown in Table \ref{tab:encoders}, PointNet outperforms the other encoders in success rates. We also freeze the point cloud encoder during the training phase of the correspondence-based diffusion policy. The results indicate that fine-tuning further enhances the learning of downstream tasks.

\begin{table}[H]
    \renewcommand{\arraystretch}{1.1} 
    \centering
    \caption{\textbf{Ablation experiments on point cloud encoders.} * indicates that the encoder is frozen during the training phase of the correspondence-based diffusion policy.}
    \label{tab:encoders}
    \footnotesize
    \resizebox{\linewidth}{!}{
        \begin{tabular}{c|cccc|c}
            \toprule
            \bf{Encoders} & \bf{PickPlace} & \bf{FlipCup} & \bf{Assembly} & \bf{ArtiManip} & \bf{Avg}\\
            \midrule
            {PointNet}
            &  {85\%} & {90\%} & {90\%} & {95\%} & {90\%} \\
            {PointNet++}
            & {5\%} & {30\%} & {30\%} & {40\%} & {26\%} \\
            {PointNeXt}
            & {0\%} & {0\%} & {0\%} & {0\%} & {0\%} \\
            {PointNet*}
            & {70\%} & {90\%} & {75\%} & {80\%} & {79\%} \\
            {PointNet++*}
            & {5\%} & {15\%} & {5\%} & {30\%} & {14\%} \\
            \bottomrule
        \end{tabular}
    }
\end{table}

\subsection{Transferability}
\label{sec: Transferability}
Our proposed CordViP can be easily transferred to different backbones. Following the design of \citet{zhao2023learning}, we choose a Transformer-based CVAE as one of the decision-making backbones for the policy, which outputs a sequence of actions. We replace the 2D image input with our Interaction-aware point clouds and use the pre-trained encoder to establish correspondences between various components. Table \ref{tab:Transferability} illustrates that CordViP achieves excellent performance on both diffusion-based and transformer-based backbones, showcasing strong transferability.

\begin{table}[H]
    \renewcommand{\arraystretch}{1.1} 
    \centering
    \caption{\textbf{Transferability to different backbone.}}
    \label{tab:Transferability}
    \footnotesize
    \resizebox{\linewidth}{!}{
        \begin{tabular}{c|cccc}
            \toprule
            \bf{Method} & \bf{PickPlace} & \bf{FlipCup} & \bf{Assembly} & \bf{ArtiManip} \\
            \midrule
            {Diffusion-based}
            &  {85\%} & {90\%} & {90\%} & {95\%} \\
            {Transformer-based}
            &  {90\%} & {75\%} & {90\%} & {80\%} \\
            \bottomrule
        \end{tabular}
    }
\end{table}

\subsection{Failure Case Analysis}
 In this section, we analyze some failure cases of CordViP. We take the flipcup task as an example and visualize the results in Figure \ref{case1}. We observe that while CordViP generally achieves good localization and grasping, the hand occasionally hovers in place after wrist rotation. One possible reason is that, during the collection of expert demonstrations, we deliberately reduced wrist speed, resulting in slight jitters that led the model to learn suboptimal solutions. A potential solution could be expanding the policy's planning horizon by increasing the values of $horizon$ and $n\_action\_steps$.

 Nevertheless, we also observe that our model possesses the ability to automatically correct some failure scenarios and recover from them. In the assembly task, inserting the cylindrical cup into the kettle is a delicate operation that relies on strong 3D spatial reasoning capabilities. As shown in Figure \ref{case2}, CordViP is able to make timely adjustments, such as when the cup is stuck at the opening of the kettle. It can quickly sense the state of the object and adjust the joint pose to successfully complete the assembly task.

\section{Conclusions and Limitations} 
\label{sec:conclusion}

In this paper, we present CordViP, a novel framework that learns correspondence-based visuomotor policy for dexterous manipulation in the real world. First, we utilize powerful 3D generation methods to obtain 3D models of objects, then use robust 6D object pose estimation based on this modeling information and robot proprioception to obtain interaction-aware point clouds, enhancing the quality of the point clouds and addressing the issues caused by occlusions from the dexterous hand. Second, we use object-centric contact map and coordination information to design a pre-training task that effectively establishes spatial and temporal correspondences. Finally, the features obtained from the pre-trained encoder are used as conditions to train a visuomotor policy.
CordViP significantly outperforms state-of-the-art 2D and 3D baselines on six real-world dexterous manipulation tasks, demonstrating highly competitive performance in both effectiveness and efficiency.

\textbf{Limitations.} Despite the exceptional performance demonstrated by CordViP, there are still certain limitations that could be explored in future work. First, our method struggles to accurately estimate the 6D pose of deformable objects due to the limited expressive capacity of FoundationPose when handling non-rigid geometry. Second, the accuracy of digital twin modeling can significantly impact pose tracking and the quality of initial object point clouds, which are critical for fine-grained dexterous manipulation. Additionally, if the dexterous hand completely occludes the object, FoundationPose may fail to track it. One promising research direction could be the incorporation of visuotactile 6D pose tracking. We leave further exploration of these possibilities for future work.

\section*{Acknowledgement}
This work was supported by the National Natural Science Foundation of China (62476011) and the Joint R\&D Fund of Beijing Smart-chip Microelectronics Technology Co., Ltd (SGSC0000RGJS2401830, SGSC0000RGJS2401829).
 

\bibliographystyle{plainnat}
\bibliography{references}

\clearpage
\appendix
\subsection{Real-World Task Description}
\label{appendix: Real-World Task Description}
\textbf{PickPlace:} 
This task requires coordinated motion of the robot’s hand and arm, involving all four fingers. The Leaphand first locates the position of the toy chicken based on visual input. The palm then approaches the chicken, with all four fingers gradually wrapping around it. Once the chicken is grasped, the wrist is raised,  and the hand is moved towards a blue bowl. Upon reaching a position directly above the bowl, the fingers are released, allowing the chicken to be placed inside. This task presents challenges in accurately locating and grasping the toy chicken, coordinating finger movements for a stable grip, and precisely manipulating the hand to place the object into the bowl while avoiding obstacles and ensuring a gentle release. Success is achieved if the toy chicken is placed into the bowl.

\textbf{FlipCup:} 
This task requires the robot to flip a cup from a lying position on the table to an upright standing position. The Leaphand approaches the cup and places its hand on the top of the cup, lifts it, and then flips it to an upright position. The hand must apply controlled force to rotate the cup while maintaining stability to prevent it from tipping over. The challenge lies in the fact that the cup may undergo changes in its orientation, requiring the hand to dynamically adjust its position and force to stabilize the cup's motion and achieve the desired outcome. Success is achieved when the cup stands upright on the table.

\textbf{Assembly:} 
This task requires the robot to assemble a cylindrical cup onto a kettle. The Leaphand first approaches the cup and grasps it, positioning it accurately to align with the kettle's opening. Using coordinated finger and wrist movements, the hand carefully attaches the cup to the kettle, ensuring a secure fit. The challenges lie in precise alignment and making fine adjustments based on feedback while handling high occlusion and ambiguity. Success is achieved when the cup is securely assembled onto the kettle.

\textbf{ArtiManip:} 
This task requires the robot to open a box using its thumb and index fingers. The robot needs to first reach the box, grasp the box's lid, and then carefully adjust its fingers to open the box without pushing it. The hand must coordinate the motion of the thumb with fine adjustments to apply the right amount of force, ensuring the lid is opened smoothly and safely. This task presents a challenge in handling articulated objects with multiple moving parts while maintaining delicate control over the manipulation process. Success is achieved when the lid is fully open.

\textbf{FlipCap:} 
This task requires the robot hand to flip over a small cap using coordinated motion of multiple fingers. The robot must first reach the cap and stabilize it with the thumb and middle finger, gently lifting and rotating it to a tilted position. Simultaneously, the index finger is required to push from the opposite side with precise force to complete the flipping motion. This task demands high-level coordination between fingers, particularly in managing the contact points during rotation. Accurate finger placement and dynamic re-adjustments are crucial to avoid slipping or excessive force. Success is defined as fully flipping the cap without displacing it.

\textbf{LongHoriManip:}
This task requires the robot hand to complete a long-horizon manipulation sequence involving four consecutive subtasks: pull, pick, place, and push. The dexterous hand must locate and grasp the drawer handle, pulling it open with sufficient force and precision. Once the drawer is fully extended, the hand proceeds to pick up a small toy chicken from the table using a coordinated grasp. It then carefully places the toy into the open drawer, ensuring stability and proper placement. Finally, the hand pushes the drawer closed, completing the sequence. This task poses a significant challenge due to its extended temporal dependencies and the need for continuous, adaptive control across multiple object interactions. Each subtask is individually evaluated for success to better understand performance at different stages. The overall task is considered successful when the toy is placed inside and the drawer is fully closed. 

We list the parameters of expert demonstrations for different tasks in Table \ref{tab:parameters of task}. For all demonstrations of a given task, we maintain a consistent number of steps. ``Demo" refers to the number of demonstrations collected for each task, ``Episode Length" denotes the duration of each episode in a task, ``Teleop. Times"  indicates the teleoperation time required to collect a single demonstration, and ``Max Steps" represents the maximum execution time for a task during evaluation.
\begin{table}[H]
    \renewcommand{\arraystretch}{1.2} 
    \centering
    \caption{\textbf{Parameters of expert demonstrations for real-world tasks.} ``Demo" refers to the number of demonstrations, ``Episode Length" denotes the duration of each episode in a task, ``Teleop. Times"  indicates the teleoperation time per demonstration, and ``Max Steps" represents the maximum execution time for a task during evaluation.}
    \label{tab:parameters of task}
    \footnotesize
    \resizebox{\linewidth}{!}{
        \begin{tabular}{c|cccc}
            \toprule
            \bf{Task Name} & \bf{Demos} & \bf{Episode Length} & \bf{Teleop. Times(s)} & \bf{Max steps} \\
            \midrule
            {\textbf{\textit{PickPlace}}}
            &  50 & 150 & 30 & 400 \\
            {\textbf{\textit{Flipcup}}}
            &  50 & 150 & 30 & 300 \\
            {\textbf{\textit{Assembly}}}
            &  50 & 175 & 35 & 500 \\
            {\textbf{\textit{ArtiManip}}}
            &  50 & 190 & 38 & 600 \\
            {\textbf{\textit{Flipcap}}}
            &  50 & 150 & 30 & 400 \\
            {\textbf{\textit{LongHoriManip}}}
            &  50 & 250 & 50 & 800 \\
            \bottomrule
        \end{tabular}
    }
    \vspace{-1em}
\end{table}

\subsection{Implementation Details}
\textbf{Network Architecture.}
For point cloud encoding, we first use PointNet\cite{qi2016pointnet} to process point cloud data without RGB information, outputting a set of point feature vectors at the dimension of 1024. The PointNet consists of three fully connected layers, each followed by LayerNorm for normalization and ReLU activation. 

For the cross-attention transformer, we adopted the architecture design from \citet{eisner2024deep}, using a multi-head attention block of 4 heads. The state features of the robotic arm and the dexterous hand are each passed through a linear layer, mapped to 16 dimensions. The features are then processed through the same Transformer architecture for cross-attention, enabling feature fusion. The fused features are subsequently combined with the original features using a residual connection.
 
\begin{figure}[H]
    \centering
    \includegraphics[width=\linewidth]{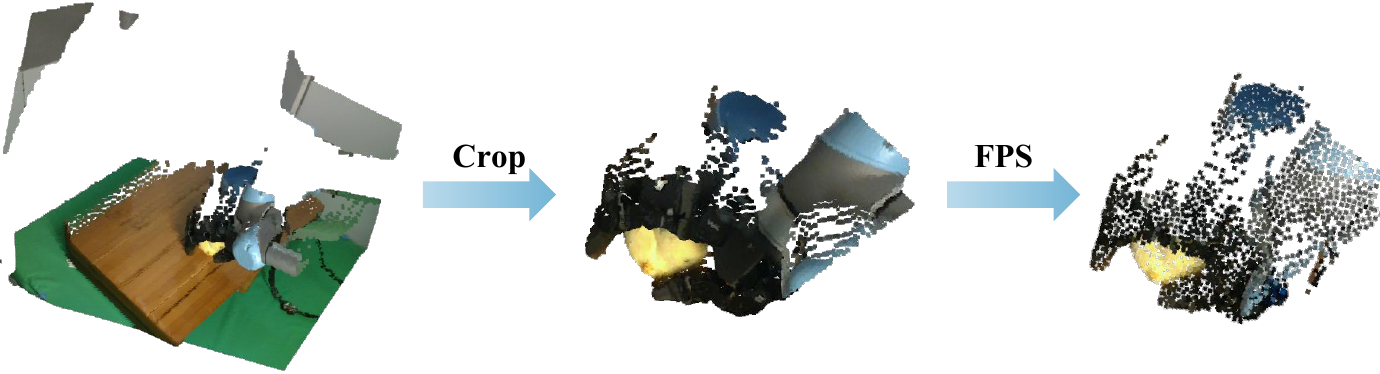}
    \caption{\textbf{Visualization of Point Cloud Processing.} The point cloud is synthesized from RGBD data. The point cloud is then cropped and processed using farthest point sampling (FPS) to generate 1024 points.}
    \label{fig:crop}
\end{figure}

\textbf{Demonstrations Process.}
We utilize the RealSense L515 camera to capture RGB-D images with a resolution of 480 $\times$ 640. The depth data are aligned with the RGB data to ensure accurate spatial correspondence. All data collection is managed through ROS and data recording begins once both the camera feed and robot teleoperation inputs are received. For our method, we use only RGB and depth data to track the object's pose. In contrast, for other baselines, we synthesize the point cloud from RGBD data, and both the pose and the point clouds are transformed into the world coordinate system. We crop point clouds with the range of $x \in [-0.4m, 0.1m], y \in [-0.7, -0.4], z \in [0.1, 0.51]$, which has been verified to be suitable for observation. as shown in Figure \ref{fig:crop}. 

We collect both the robot's state and actions using joint angles in radians, including the 6-DOF joints of the robotic arm and the 16-DOF joints of the Leaphand.

\begin{figure}
    \centering
    \includegraphics[width=\linewidth]{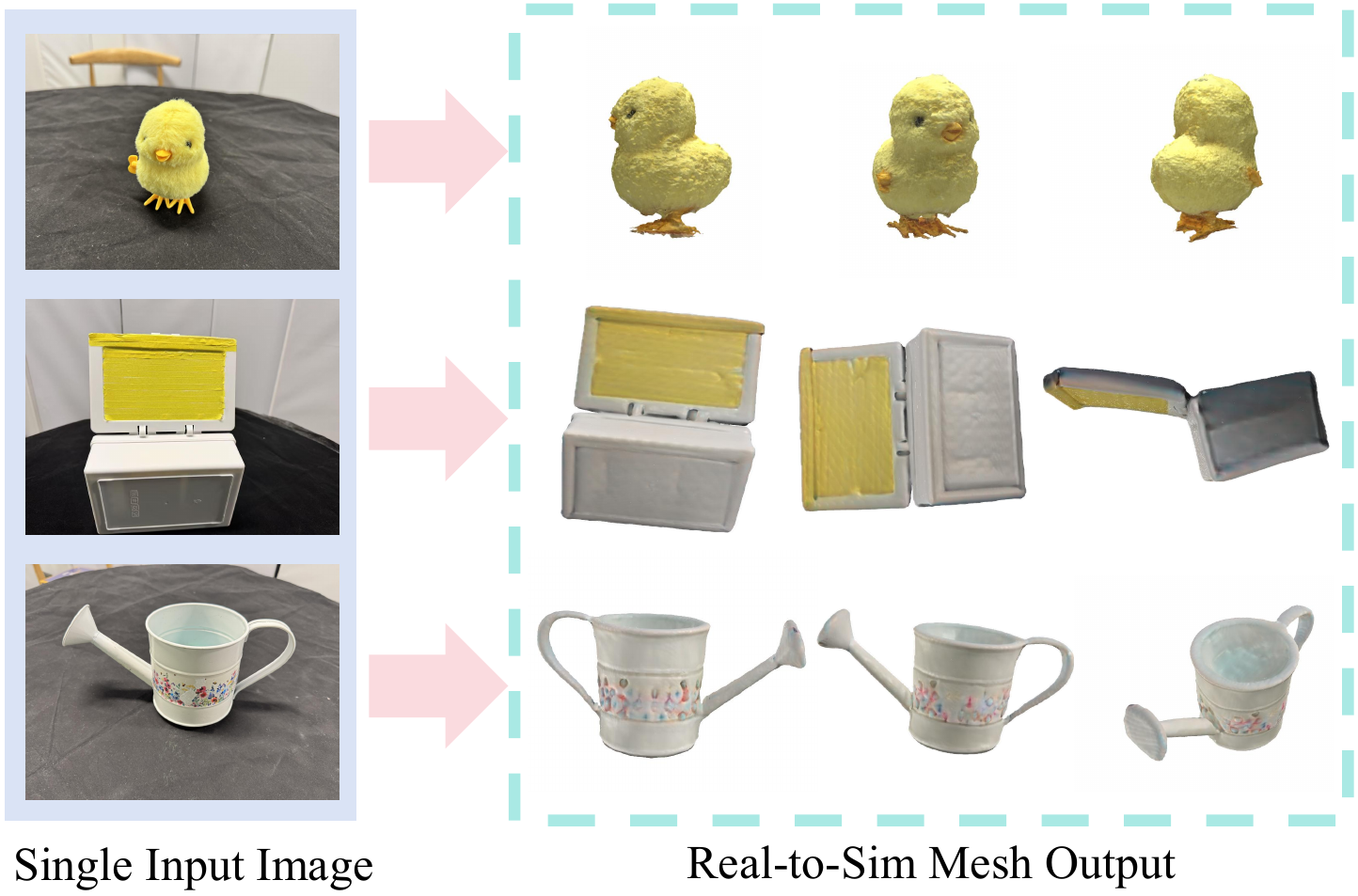}
    \caption{3D digital asset generation from a single view.}
    \label{fig:real2sim}
\end{figure}

\textbf{Real-to-Sim for Digital Twin Generation.} We use TripoSR \cite{TripoSR2024} to generate digital twins from a single-view image, which enables the creation of high-quality 3D assets. The visual results are shown in the figure \ref{fig:real2sim}.

\textbf{6D Pose Estimation.} We utilize FoundationPose \cite{foundationposewen2024} to perform robust 6D pose estimation for various objects across tasks. For the PickPlace task, we estimate the 6D pose of both the chicken and the bowl to capture the spatial relationships between the objects. For the FlipCup task, we focus on accurately estimating the 6D pose of the cup. For the Assembly task, we separately estimate the 6D pose of the cylindrical cup and the kettle, enabling precise interactions with each object during assembly. For the Artimaip task, we first decompose it into two distinct parts: the box body and the lid, and then perform estimation for each part using FoundationPose. For the FlipCap task, we estimate the 6D pose of the cap. For the LongHoriManip task, we separately estimate the 6D poses of the toy chicken and the drawer handle.

\textbf{Normalizations.}
The range of training data has a significant impact on the training stability of CordViP. We linearly scale the minimum and maximum values of each action and observation dimension to the range of [-1, 1]. This step is necessary for DDIM \cite{song2020denoising} and DDPM \cite{ho2020denoising}, as they clip the predicted results to the range of [-1, 1] for training stability.

\textbf{Hyperparameters.}
We list the training hyperparameters used in CordViP in Table \ref{tab:hyperparameters}.

\begin{table}[H]
    \renewcommand{\arraystretch}{1.1} 
    \centering
    \caption{Training hyperparameters in CordViP.}
    \label{tab:hyperparameters}
    \footnotesize
        \begin{tabular}{cc}
            \toprule
            \bf{Hyperparameters} & \bf{Value}\\
            \midrule
            {Robot point cloud size} &  {1024*3}  \\
            {Object point cloud size} & {1024*3}  \\
            {Contact map scaling factor $\gamma$} & {1} \\
            {Contact map scaling factor $\theta$} & 10 \\
            {Contact map size} & {1024*1} \\
            {Loss weight $\lambda$} & {1} \\
            {$horizon$} & {12} \\
            {$n\_obs\_steps$} & {4} \\
            {$n\_action\_steps$} & {6} \\
            {Optimizer} & {AdamW} \\
            \bottomrule
        \end{tabular}
    \vspace{-1em}
\end{table}

\textbf{Baselines settings.}
We train the Diffusion Policy for 600 epochs with $horizon$=12, $n\_obs\_steps$=4, and $n\_action\_steps$=8. The Diffusion Policy baseline utilizes ResNet18 as the visual encoder and employs CNN-based backbones.
The 3D Diffusion Policy is trained for 8000 epochs with $horizon$=12, $n\_obs\_steps$=4, $n\_action\_steps$=8. It uses DP3 Encoder as the point cloud encoder. 
For ACT, we train the model for 5000 epochs with $action\_chunk$=30. 
In ACT3D, train for 1600 epochs with an $action\_chunk$ size of 30. Point clouds are used to replace the original image inputs. The point clouds are encoded using PointNet, and the extracted point cloud features are given learnable positional encodings, similar to other input information such as joint states and latent inputs. 
For BCRNN, we train the model for 1500 epochs with $horizon$=10, $n\_obs\_steps$=1, $n\_action\_steps$=1. 
The BCRNN3D is trained for 3000 epochs with $horizon$=10, $n\_obs\_steps$=1, $n\_action\_steps$=1, where the observations are replaced from images to point clouds. It uses PointNet as the point cloud encoder.

\subsection{Failure Analysis of Baselines}
\label{appendix: Failure Case Analysis}
\textbf{3D Diffusion Policy.} DP3 \cite{ze20243d} appears to struggle in learning meaningful actions from our demonstration data. We have already analyzed in the paper that potential reasons include factors such as the camera's viewpoint and the quality of the point cloud. \citet{wang2024rise} also points out that the type of motion patterns can affect the quality of demonstration learning. Instead of nature actions, axis-wise actions were used in DP3's demonstration data. This is because the robotic arm is controlled by the keyboard, which inherently limits the motion representation to axis-wise actions.
\begin{figure}[H]
    \centering
    \includegraphics[width=0.9\linewidth]{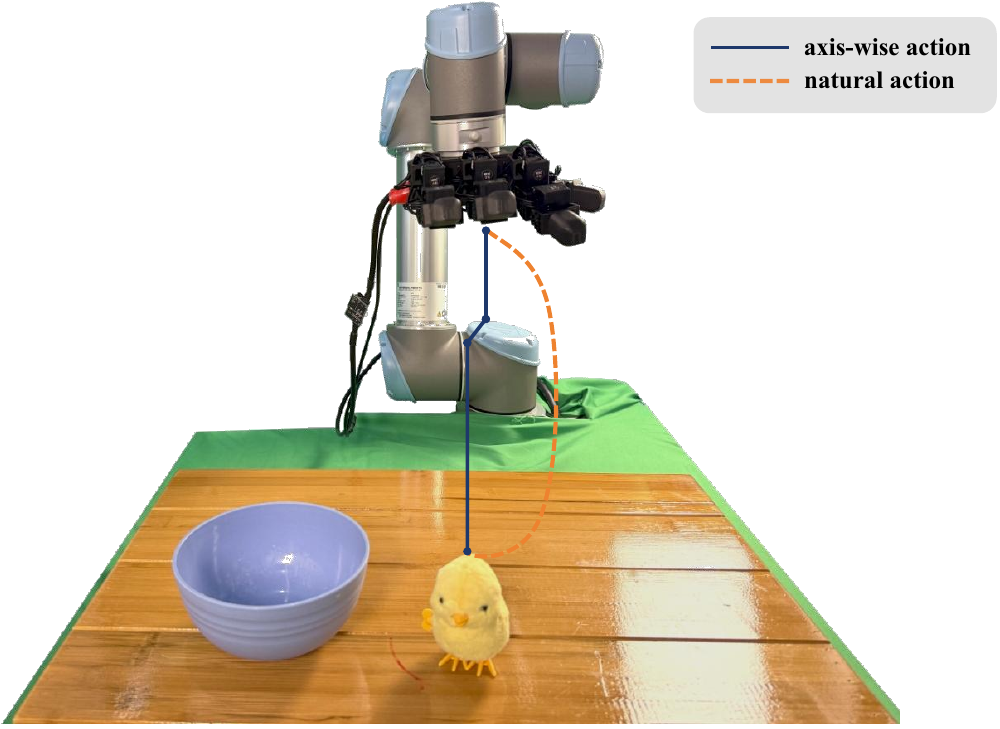}
    \caption{\textbf{Comparison of Motion Patterns.} DP3 Uses Axis-Wise Actions.}
    \label{fig:motion patten}
\end{figure}

\begin{figure*}[h]
    \centering
    \includegraphics[width=\linewidth]{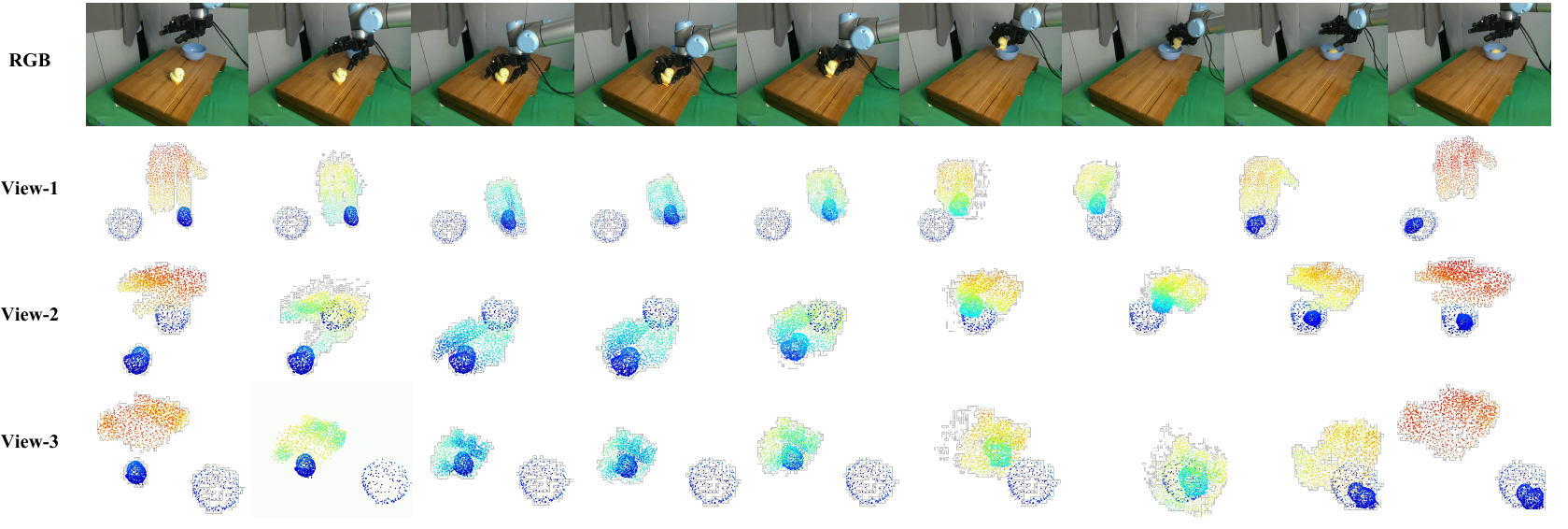}
    \includegraphics[width=\linewidth]{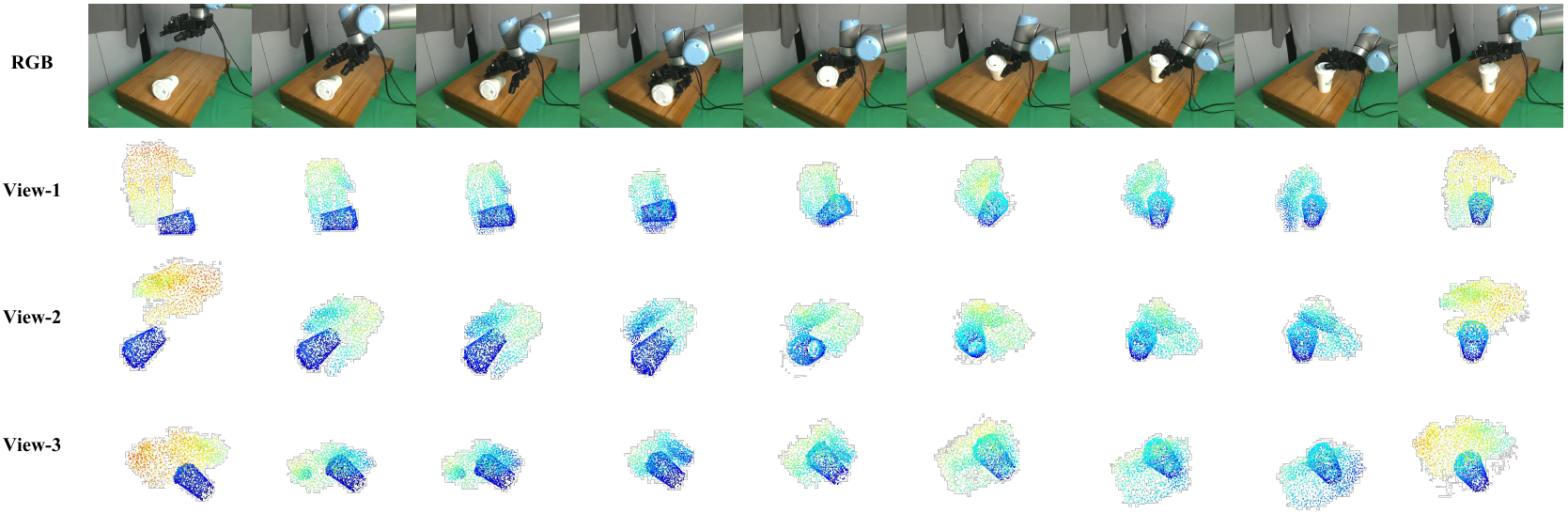}
    \includegraphics[width=\linewidth]{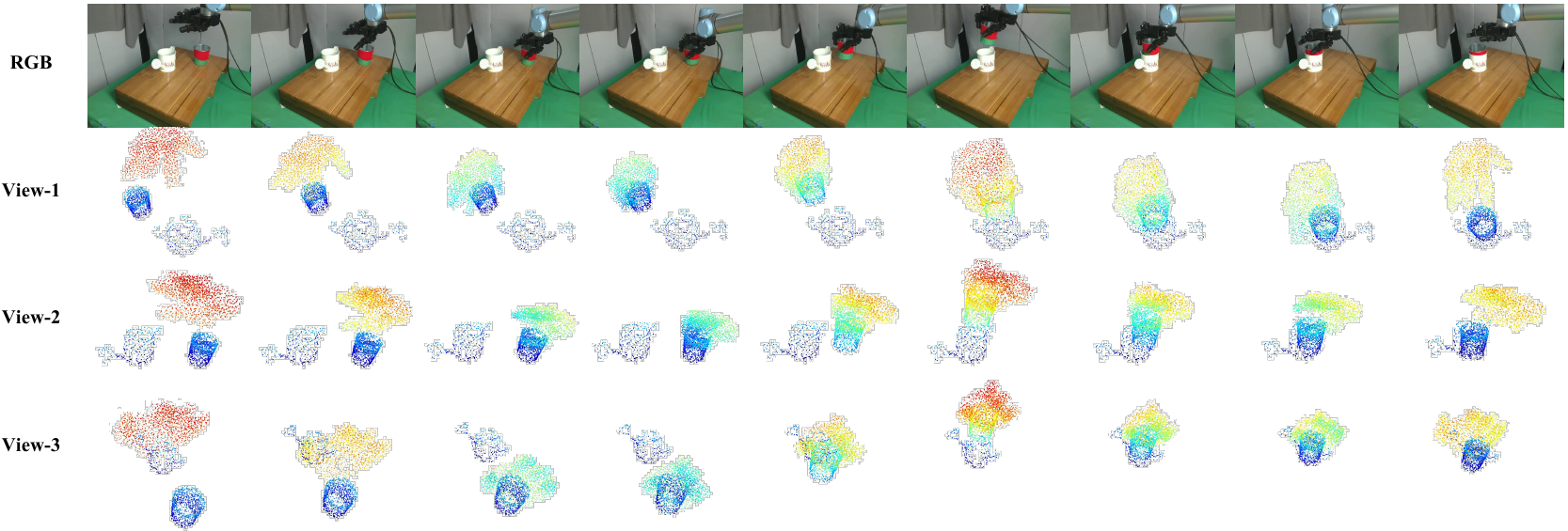}
    \caption{\textbf{Visualization of Point Clouds During the Task Process.}}
    \label{fig:vis_pc}
\end{figure*}

\textbf{RISE.} RISE \cite{wang2024rise} is a recently proposed end-to-end baseline for real-world imitation learning, which predicts continuous actions directly from single-view point clouds. It takes voxelized point clouds as input to the policy and assumes that the end-effector pose of the robotic arm is implicitly encoded within the point cloud. However, this approach proves unsuitable for our dexterous hand scenario, where the hand has a high degree of freedom and often experiences occlusions. We evaluated RISE in our settings and observed that the robotic arm exhibited excessively abrupt movements.

\subsection{More Visualization Results}
We present additional point cloud visualization results in Figure \ref{fig:vis_pc}, demonstrating that interaction-aware point clouds can effectively enhance the quality of 3D observations.

\end{document}